\pdfoutput=1
\documentclass[11pt]{article}

\usepackage[final]{acl}

\usepackage{times}
\usepackage{latexsym}
\usepackage[T1]{fontenc}
\usepackage[utf8]{inputenc}

\usepackage{inconsolata}
\usepackage{etoolbox}

\usepackage{graphicx}

\newcommand{\mymethod}{\textsc{Rationalyst}}

\usepackage[utf8]{inputenc} % allow utf-8 input
\usepackage[T1]{fontenc}    % use 8-bit T1 fonts
\usepackage{hyperref}       % hyperlinks
\usepackage{url}            % simple URL typesetting
\usepackage{booktabs}       % professional-quality tables
\usepackage{amsfonts}       % blackboard math symbols
\usepackage{nicefrac}       % compact symbols for 1/2, etc.
\usepackage{xcolor}         % colors
\usepackage{graphicx} 
\usepackage{amsmath}
\usepackage{multirow}

\usepackage{float}

\usepackage{placeins}
\usepackage{graphicx}

\usepackage{tikz}

\newcommand*\circled[1]{\tikz[baseline=(char.base)]{
            \node[shape=circle,draw,inner sep=0.2pt] (char) {#1};}}

\usepackage{etoolbox}
\robustify{\circled}

\usepackage{array}
\newcommand{\PreserveBackslash}[1]{\let\temp=\\#1\let\\=\temp}
\newcolumntype{C}[1]{>{\PreserveBackslash\centering}p{#1}}
\newcolumntype{R}[1]{>{\PreserveBackslash\raggedleft}p{#1}}
\newcolumntype{L}[1]{>{\PreserveBackslash\raggedright}p{#1}}
\usepackage{color,xcolor}
\usepackage{epsfig}
\usepackage{graphicx}

\usepackage{adjustbox}
\usepackage{array}
\usepackage{booktabs}
\usepackage{colortbl}
\usepackage{float,wrapfig}
\usepackage{hhline}
\usepackage{multirow}
\usepackage{subcaption} %
\usepackage[font={small}]{caption}
\usepackage{natbib}
\usepackage{comment}
\usepackage{amsmath,amsfonts,amsthm,amssymb}
\usepackage{bm}
\usepackage{nicefrac}
\usepackage{microtype}
\usepackage{mathtools}

\usepackage{changepage}
\usepackage{extramarks}
\usepackage{fancyhdr}
\usepackage{lastpage}
\usepackage{setspace}
\usepackage{soul}
\usepackage{xspace}

\usepackage{amsmath}
\usepackage{url}
\usepackage{xpatch}
\usepackage{algorithm}
\usepackage{algpseudocode}
\usepackage{todonotes} %
\usepackage{enumitem}
\usepackage{titlesec}
\usepackage{bbm}

\newcommand{\logo}[0]{\includegraphics[width=.034\textwidth,trim=0cm 0.99cm 0cm 0cm]{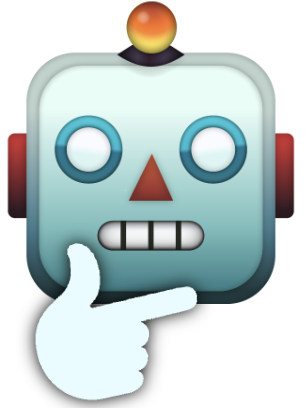}}

\title{
\vspace*{-0.5in}
{{\small \hfill ACL'25}\\
\vspace*{.25in}} 
\logo~\mymethod{}: 
Mining Implicit Rationales \\for Process Supervision of Reasoning}
\author{Dongwei Jiang$^{\spadesuit}$ \quad Guoxuan Wang$^{\spadesuit}$ \quad Yining Lu$^{\diamondsuit}$ \quad Andrew Wang$^{\spadesuit}$\\
\textbf{Jingyu Zhang}$^{\spadesuit}$ \quad \textbf{Chuyu Liu}$^{\spadesuit}$ \quad \textbf{Benjamin Van Durme}$^{\spadesuit}$ \quad \textbf{Daniel Khashabi}$^{\spadesuit}$
\\
$^{\spadesuit}$Johns Hopkins University, $^{\diamondsuit}$University of Notre Dame\\
\url{djiang21@jhu.edu}
}

\definecolor{darkred}{RGB}{200,0,0}
\definecolor{lightgreen}{RGB}{231,255,219}
\definecolor{lightred}{RGB}{252,231,234}
\definecolor{lightyellow}{RGB}{250,253,191}
\definecolor{DarkRed}{RGB}{130,25,0}
\definecolor{purplebg}{RGB}{229, 199, 244}

\begin{document}

\maketitle

\begin{abstract}

The reasoning steps generated by LLMs might be incomplete, as they mimic logical leaps common in everyday communication found in their pre-training data: underlying rationales are frequently left \emph{implicit} (unstated). To address this challenge, 
% we introduce \mymethod{}, 
% \st{a model for process-supervision of reasoning based on pre-training on a vast collection of rationale annotations extracted from unlabeled data.} 
we introduce \mymethod{}, a three-stage approach for improving LLM reasoning: (1) extracting implicit rationales from unlabeled data, (2) training a specialized rationale generation model, and (3) using these rationales to provide process supervision during inference.
We extract 79k rationales from web-scale unlabelled dataset (the Pile) and a combination of reasoning datasets with minimal human intervention. This web-scale pre-training for reasoning allows \mymethod{} to consistently generalize across diverse reasoning tasks, including mathematical, commonsense, scientific, and logical reasoning.
Fine-tuned from LLaMa-3-8B-Instruct, \mymethod{} improves the accuracy of reasoning by an average of 3.9\% on 7 representative reasoning benchmarks. It also demonstrates superior performance compared to significantly larger verifiers like GPT-4 and similarly sized models fine-tuned on matching training sets. \footnote{
 Our code, data, and model can be found 
 at this 
 repository: 
\url{https://github.com/JHU-CLSP/RATIONALYST}
 }

%Our approach involves three stages. First, we use large language models (LLMs) to distill implicit rationales from unlabelled text corpora, uncovering the underlying explanations or justifications for conclusions and inferences. Those rationales are subsequently filtered based on whether they help predict the following text. Next, we train \mymethod{} to predict these rationales given the preceding context. Finally, we utilize the predicted rationales to develop heuristics that supervise the reasoning process at inference time.
%
%Our approach provides process supervision through rationales that 
% These rationales, extracted from various reasoning datasets as well as web-scale unlabeled data, are straightforward for humans to understand while also providing process supervision for models during reasoning.
% \daniel{highlight: 
% (1) size of pre-training data for your extraction, or the size of the extracted/filtered data you used for training; 
% (2) number of datasets you evaluate; 
% }

\end{abstract}

\section{Introduction}

Rationales play a crucial role in human reasoning and its accuracy \citep{rips1994psychology, mercier2011humans}. In reasoning problems, having accurate rationales often correlates with accurate outcomes \citep{tversky1982judgment, davis1984diagnostic}. This importance of rationales extends to Large Language Models (LLMs) as well. \citet{wei2022chain} were among the first to show that generating chain-of-thought rationales significantly improves LLMs' reasoning performance. Subsequent research has further refined the methods for eliciting rationales, leading to improved performance \citep{ zhou2022reflectionthoughtinverselyeliciting}.

\newcommand{\mycolors}[1]{\colorbox{mygray}{\textcolor{white}{#1}}}

% Reduce the padding in the colorbox
\setlength{\fboxsep}{1.0pt} % Change this value as needed

\begin{figure}[t] % Use [t] for top, [b] for bottom, or [h] for here
    \centering
    \includegraphics[trim=2.46cm 2.0cm 12.4cm 0cm, scale=0.55]{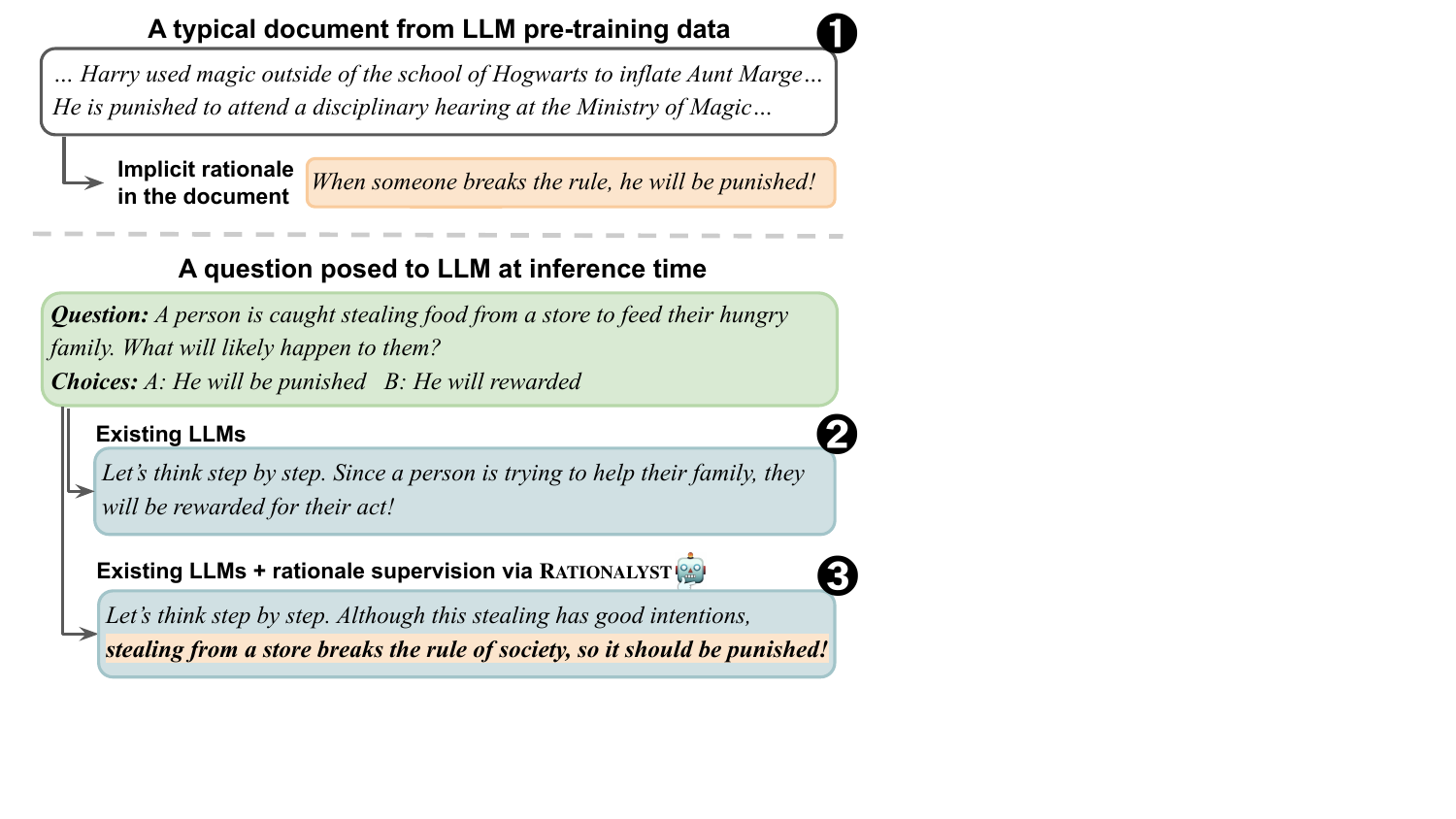}
    \caption{\textbf{A simplified example showing how implicit rationales in pre-training data can be leveraged to improve reasoning.} \circled{1}: Implicit rationales (unstated logical connections) occur frequently in LLM pre-training data. \circled{2}: Existing LLMs pre-trained to replicate their pretraining data tend to omit these logical steps as well. \circled{3}: However, \mymethod{} learns to generate these rationales at inference time to supervise the chain-of-thought process for better reasoning.}
    % An example of how implicit rationales extracted from LLM pre-training data can guide inference-time reasoning. Top: Implicit rationales exist in everyday text. Bottom: RATIONALYST generates these rationales during inference, improving the accuracy of reasoning.
    \label{fig:teaser}
\end{figure}

In the context of LLM reasoning, these rationales are typically employed through a chain-of-thought process that makes reasoning \emph{explicit} by articulating them as plain-text rationales.
In this approach, each subsequent rationale is generated conditioned on rationales produced in preceding steps, effectively using them as a form of supervision. 
%\dongweicomment{will this deviate the point of the paper? However, this approach faces several challenges. First, it requires a significant amount of human effort to create demonstrations that elicit these rationales. Second, these explicit rationale demonstrations are typically tailored to specific problems, making it difficult for models to generalize.} 
However, 
% when faced with new problems, 
the generated reasoning chains might be incomplete, containing potential logical leaps while leaving some rationales \emph{implicit} (or hidden) during the generation process. These gaps in the reasoning chain can weaken the LLM's reasoning ability throughout the problem-solving process.

% However, this approach presents three challenges: first, a significant amount of human labor is required to generate demonstrations that elicit these rationales; second, explicit rationale demonstrations are typically designed for specific problems, making it difficult for models to generalize. \dongweicomment{is this motivation good?} 
% \st{Previous research has primarily focused on generating rationales \emph{explicitly} within answers through a chain-of-thought process. In this approach, each subsequent rationale is generated based on rationales produced in preceding steps, effectively using them as a form of supervision. However, the reasoning chains produced in this manner might be incomplete. 
% % \dongweicomment{is this the best way to motivate the problem?} 
% This potential incompleteness stems from intuitive logical leaps where some rationales remain \emph{implicit} (or hidden) during the generation process. These possible gaps in the reasoning chain can make it challenging to provide consistent and comprehensive supervision throughout the problem-solving process.}
\begin{figure*}[t]
  \centering
  % Adjust widths to fit your document
 
    \centering
    \includegraphics[trim=0.0cm 7.8cm 5.5cm 0cm, width=0.98\linewidth]{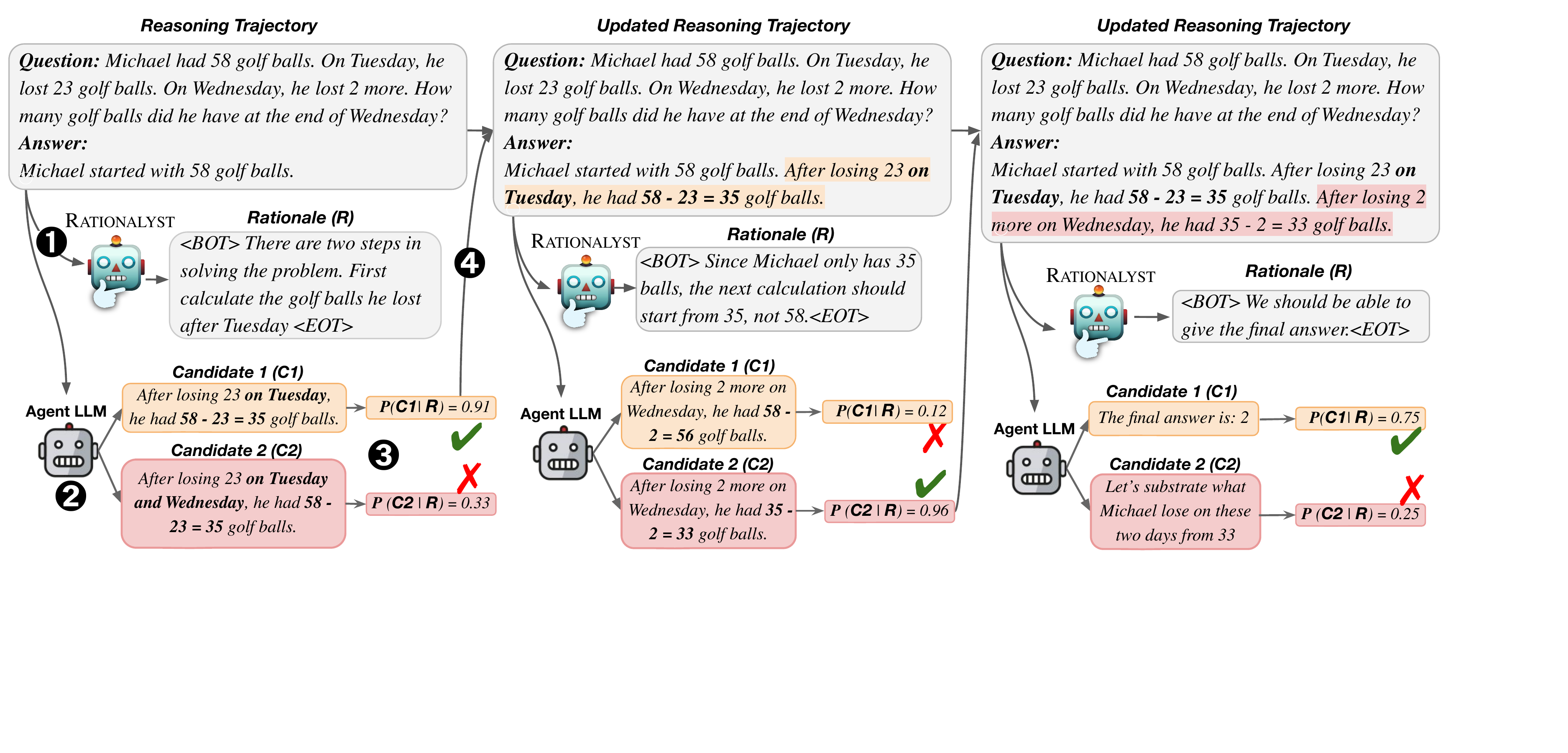} 
    \caption{\textbf{An example showing how \mymethod{} works at inference time.} \mymethod{} generates implicit rationales given the current reasoning trajectory, which includes both the question and the reasoning steps generated so far \circled{1}. Agent LLM generates multiple next-step candidates for reasoning, also based on the current reasoning trajectory \circled{2}. Implicit rationale generated by \mymethod{} is used to provide heuristics for choosing the next step candidates proposed by the agent LLM by estimating the probability of the next step candidate given the rationale \circled{3}. The reasoning trajectory is updated iteratively with the highest scoring next step candidate \circled{4}.}
    \label{fig:main_model_step_3}
\end{figure*}

% In this example, implicit rationales can be defined as rationales that are inherent but not explicitly articulated. They are common also in unlabelled pre-training data.

One reason why chain-of-thought methods might miss implicit steps is that models trained with ``next-token prediction''  often replicate the omissions present in their training data. Implicit rationales--underlying logical connections that are often not explicitly stated--are frequently missing in daily communication and web text.
\autoref{fig:teaser} \circled{1}
% (top) 
illustrates this concept using a typical document from LLM pre-training data. In this example, we see a passage from Harry Potter: ``Harry used magic outside... He is punished to attend..." The text contains the implicit (unstated) rationale: \emph{``When someone breaks the rule, he will be punished!"} 
This implicit rationale is crucial in inferring the causal reasoning that connects the cause (\emph{Harry breaking rules}) to its
effect (\emph{punishment}), but is also left unstated in the context.
As a result, existing LLMs trained to mimic web text will have difficulty surfacing these implicit statements during the reasoning process, which can lead to flawed conclusions, such as erroneously justifying theft as a praiseworthy act when done to support one's family (\circled{2} in \autoref{fig:teaser}).

This paper presents \mymethod{}, a model tailored for process-supervision of reasoning. \mymethod{} is trained on a vast collection of implicit rationales extracted from a mixture of web-scale unlabeled datasets and existing reasoning datasets. 
Although existing LLMs may miss crucial details in their reasoning, leading to flawed conclusions (\circled{2} in \autoref{fig:teaser}), using \mymethod{} provides additional supervision mechanism to guide their reasoning processes, resulting in more robust conclusions (\circled{3} in \autoref{fig:teaser}).

% This paper presents \mymethod{}, a model designed to address the challenge of extracting and utilizing implicit rationales for reasoning. 
% As shown in  the lower portion of Figure \ref{fig:teaser}, LLMs may overlook crucial unstated principles without the supervision of implicit rationales. This hypothetical scenario demonstrates how such an oversight can lead to flawed conclusions, such as erroneously justifying theft as a praiseworthy act when done to support one's family. 
% In contrast, \mymethod{} learns implicit rationales from unlabelled data and utilizes them to guide reasoning with greater accuracy. \dongweicomment{let me know if the wording here is still weird}

\mymethod{} is developed and used in three stages: 
(1) we employ LLMs to extract implicit rationales from unlabeled text corpora without human annotation. These rationales are subsequently filtered based on their helpfulness in predicting subsequent text (\S\ref{sec:rationale_distillation}); (2) we train \mymethod{} to predict those rationales given the preceding context (\S\ref{sec:world_model_training});
and then (3) as depicted in  \autoref{fig:main_model_step_3}, during inference, we assume reasoning is done incrementally in a chain-of-thought fashion \citep{wei2022chain} by another agent model, and we use \mymethod{} to provide supervision for the agent model at each reasoning step throughout the reasoning process \S\ref{sec:world_model_inference}. 
By adopting a data-centric approach, \mymethod{} utilizes abundant unlabelled data to provide process supervision \citep{lightman2023lets} across various reasoning tasks without the need for human annotation. 
% \dongweicomment{talk more about advantage?}\daniel{yes}
% \dongweicomment{put figure 3 to page 2 and refer to this figure only, figure 2 can be pushed down}

% The power of such implicit rationales becomes evident when applying them to new, similar scenarios. Going back to Figure \ref{fig:teaser}, when asked about the likely outcome for a person caught stealing from a store, we can leverage the same implicit rationale (which can be generated by \mymethod{}). By recognizing that stealing also breaks rules, we can infer that the person will likely be punished rather than rewarded.
% Employing a data-centric approach, \mymethod{} makes use of the abundant pre-training data available to us while providing process supervision \citep{lightman2023lets} that is general for various reasoning tasks. \dongweicomment{Notably, this supervision is achieved without the need for human annotation.}
% More details can be found in \S\ref{sec:method}. 
% \dongweicomment{talk more about advantage?}\daniel{yes}

Our method extracts 65k implicit rationales from the web-scale unlabelled dataset The Pile \citep{gao2020pile}. To adapt the extracted rationales to our tested domain and stabilize training, we additionally extract a much smaller set of 14k implicit rationales from the question-answer pairs in the training sets of two reasoning datasets: GSM8K and ECQA. 
Our extraction process controls for answer leakage to prevent artificial amplification of performance. Using this curated set of rationales, \mymethod{} is then fine-tuned from LLaMa-3-8B.
To assess the effectiveness of our approach, we evaluate \mymethod{} on a diverse set of reasoning tasks, including mathematical, commonsense, scientific, and logical reasoning. Our results show that \mymethod{} improves the accuracy of reasoning by an average of 3.9\% (\S\ref{sec:big_world_model}). To understand the contribution of different data sources, we conduct an ablation study that demonstrates the utility of rationales from both the large-scale Pile dataset and the smaller, specialized reasoning datasets (\S\ref{sec:ablation_web_scale_rationales}). Notably, \mymethod{} exhibits superior performance when compared to strong general-purpose verifiers like GPT-4 and similar capacity models specifically fine-tuned on matching training sets (\S\ref{sec:comparsion_with_verfiers}).

Implicit rationales generated by \mymethod{} are also designed to provide supervision in a human-readable form, offering improved interpretability for LLM generation. This added interpretability is particularly beneficial when reasoning over complex domains such as mathematics or coding, where the step-by-step logic can be difficult for humans to follow without explicit explanations.
As shown in \S\ref{sec:intepreability}, our model is capable of generating human-understandable rationales for unseen data from complex math reasoning. Furthermore, our analysis demonstrates that \mymethod{} introduces minimal computational overhead during inference (\S\ref{sec:compute_analysis}).

% Our contributions in this paper are two-fold:
% \begin{itemize}[noitemsep,topsep=0pt,parsep=0pt,partopsep=0pt,leftmargin=10pt]
%     \item We propose \mymethod{}, a model that is pre-trained on implicit rationales extracted from unlabeled text data. \mymethod{} enhances LLM interpretability and performance during reasoning by providing process supervision.
%     \item We empirically show \mymethod{} generalizes across reasoning tasks and scales with unlabelled data.
%     %, for which we offer an empirical explanation
% \end{itemize}

\section{Related Work}
% \paragraph{Building world model.} The concept of world model has an extensive background. As described in various studies \citep{hu2023language}, world models can typically be formulated as state transition probabilities, which characterize a generative, casual mechanism of how the world state changes after an agent’s actions. 

% In the field of vision, pioneering efforts have been made in developing world models. \citet{du2023video} uses vision-language models to replicate the fundamental dynamics of the world. This world model is then used in various downstream tasks like planning  \citep{du2023video}, representation learning \citep{Garrido2024LearningAL} and robotic manipulation \citep{Bharadhwaj2024Track2ActPP}. The success of this method relies on converting diverse data formats into a uniform format \citep{yang2024learning} and utilizing generative modeling on this standardized data.

% \paragraph{Distilling hidden knowledge in pre-training corpus.} 

\paragraph{Supervising reasoning.}
Supervision-based approaches have been shown to enhance the reasoning abilities of LLMs. \citet{gsm8k} and \citet{snell2024scalingllmtesttimecompute} demonstrate that training a ``verifier" to supervise reasoning can be more parameter-efficient than simply expanding the parameters of the ``reasoner" responsible for solving the reasoning task. Ground-truth feedback from interaction with the environment is an effective form of supervision \citep{wang2023voyager}, but it works only in controlled environments like simulated world. General-purpose verifiers \citep{dhuliawala2023chainofverification, weir2024enhancing, weir2023nellie, vacareanu2024generalpurposeverificationchain} offer broader applicability utilizing principles like compositional reasoning. However, they don't fully capitalize on the vast amount of unlabelled data in the way a data-driven approach might. Process-based supervision \citep{lightman2023lets} offers supervision at each reasoning step rather than just at the final result. While promising, it requires substantial human annotation for the correctness of intermediate steps, making it resource-intensive. Our work aims to address these challenges by proposing a data-centric process-supervision method without the need for human annotation.

\paragraph{Knowledge extraction from unlabelled data.}
% next-token prediction does not always provide the complete set of reasons and often involves "leaps of logic" and hence, "next-token" prediction is not harness the full ability of LLMs to reason.
% \daniel{"potential" for what? The first sentence needs to tie to "knowledge extraction", which is the focus of this paragraph. Perhaps yo can rewrite the first 3 sentences a bit more concisely. }. 
LLMs are conventionally trained on extensive web data using autoregressive next-token prediction. While effective, this approach may not fully harness the potential of the pre-training data, as latent information within this data could be better accessed using techniques beyond simple next-token prediction.
Recent research has demonstrated several approaches to utilize this latent information to develop more sophisticated language model capabilities \citep{jiang-etal-2024-rora}.
% \daniel{"potential" for what? The first sentence needs to tie to "knowledge extraction", which is the focus of this paragraph. Perhaps yo can rewrite the first 3 sentences a bit more concisely. }. 
 \citet{schick2023toolformer} introduced Toolformer, which autonomously annotates and extracts appropriate positions, names, and inputs for tool use by leveraging supervision from future tokens. Similarly, \citet{cornille2024learningplanlanguagemodeling} developed a method for learning to plan coherent article writing through self-supervised learning in text. More closely related to our work, \citet{zelikman2024quietstar} proposed Quiet-Star, which applied a comparable technique to uncover underlying rationales in daily communication to enhance reasoning capabilities.
Our work adopts a strategy similar to Quiet-Star for extracting rationales in an unsupervised manner. However, our approach diverges in its primary objective: we aim to train a ``supervisor" that can utilize these rationales to provide process supervision for any ``reasoner." This focus enables us to implement a simpler and more reliable method, as we don't need to directly integrate rationale extraction with ``reasoner" training. Our approach thus offers a novel perspective on leveraging latent information in language models to enhance their capabilities.

\paragraph{Rationales as the basis for reasoning.} 

% Rationales play a crucial role in human reasoning and its accuracy \citep{rips1994psychology, mercier2011humans}. Correct rationales typically lead to accurate reasoning outcomes, making them useful for supervising reasoning processes \citep{tversky1982judgment, davis1984diagnostic}. For LLMs, rationales are also instrumental to their reasoning capability.
Various studies have focused on improving the use of rationales to elicit reasoning. \citet{complexity} refine rationales for more effective reasoning elicitation, while \citet{li2023making} explore different approaches to leveraging rationales to enhance reasoning. Other works, such as \citet{hwang2024selfexploreavoidpitimproving}, examine the verification of rationales produced by LLMs during reasoning to improve performance. Additionally, training LLMs on rationale-rich data is a common strategy for enhancing reasoning skills. As highlighted by \citet{quantitive_llm} and \citet{jiang2024leanreasoner}, LLMs trained on science and math data tend to perform better on reasoning tasks, particularly when CoT prompting is used. In this work, we build on this foundation by using rationales as the core of our method to supervise reasoning.
%Recent studies further discovered training on rationals can enhance LLM's reasoning capability in a general way. As noted by \citet{quantitive_llm}, LLMs trained with science and math data do better on tasks that require reasoning, especially when using CoT prompting. Other results by \citet{complexity} suggest that powerful LLMs obtain advanced reasoning capabilities from being trained on code. 
% Other results by \citet{fu2022gptroadmap} and \citet{complexity} suggest that powerful LLMs gain advanced reasoning capabilities from being trained on code, as it includes many rationales in the form of comments. 

% \paragraph{Enhancing the reasoning capability of LLMs.} Reasoning is the bedrock of human intelligence. There have been various efforts to improve reasoning performance in LLMs. Some approaches focus on better prompt engineering to aid reasoning \cite{Fu2022ComplexityBasedPF, zou2024generalizablechainofthoughtpromptingmixedtask}, while others introduce topological variants \citep{yao2023tree, jiang2024respromptresidualconnectionprompting}. Additional methods include verification and refinement \citep{paul-etal-2024-refiner}, question decomposition \citep{zhou2022least}, and knowledge enhancement \citep{dhuliawala2023chainofverification}.
% Despite these advancements, LLMs continue to face challenges, including struggles with generalization and handling complex reasoning tasks \citep{greedy_reasoner, rein2023gpqagraduatelevelgoogleproofqa, nezhurina2024alicewonderlandsimpletasks}.

\section{Building \mymethod{}}\label{sec:method}

\definecolor{mina-blue}{rgb}{0.3333, 0.3764, 0.6627}
\definecolor{mina-green}{rgb}{0.4078, 0.6392, 0.4274}
\definecolor{mina-yellow}{rgb}{0.9803, 0.7372, 0.0352}
\definecolor{mina-red}{rgb}{0.9450, 0.1, 0.1}

\newcommand{\base}{\textcolor{mina-red}{$M$}}
\newcommand{\esti}{\textcolor{mina-red}{$M$}}
\newcommand{\rationalyst}{\textcolor{mina-yellow}{$M_\text{Ra}$}}
\newcommand{\agent}{\textcolor{mina-blue}{$M_\text{Agent}$}}

\begin{figure*}[t]
  \centering
  % Adjust widths to fit your document
 
    \centering
    \includegraphics[trim=2.4cm 11cm 3cm 0.1cm, width=0.90\linewidth]{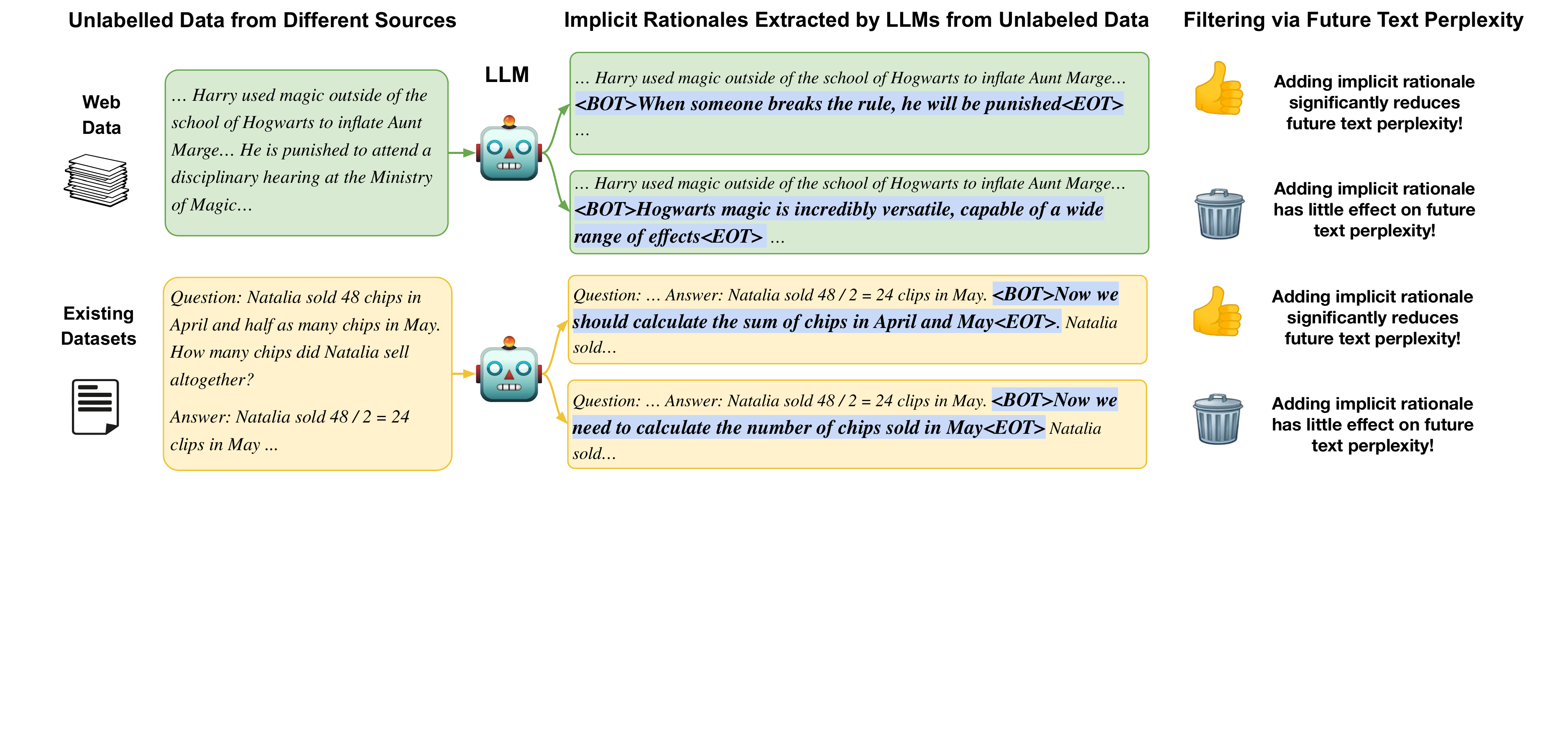} % Second figure file name
    \caption{We use LLMs to extract implicit rationales (enclosed by {\tt<BOT>} and {\tt<EOT>} in bold) that capture reasoning in unlabelled text  (\S\ref{sec:rationale_distillation} \circled{A} and \circled{B}).
The sample at the top is taken from unlabelled web-scale pre-training
datasets The Pile and the sample at the bottom is taken from existing datasets (GSM8K). These rationales are subsequently filtered based on whether they are useful for predicting future text (\S\ref{sec:rationale_distillation} \circled{C}).}%\textbf{The three stages of our method.}
    \label{fig:main_model_step_1}
\end{figure*}

We discuss the construction of \mymethod{} and its usage at inference time. 
First, we describe extracting rationales from unlabeled text (\S\ref{sec:rationale_distillation}), then use them to train \mymethod{} (\S\ref{sec:world_model_training}), and finally, employ \mymethod{} to supervise reasoning during inference (\S\ref{sec:world_model_inference}). 

\paragraph{Setup.}
As we will be using multiple LLMs throughout the process, we define them here: \rationalyst{} is the trained rationale generation model (\mymethod{}) that generates rationales and heuristics during inference. \agent{} is a general-purpose reasoning agent that produces candidate reasoning steps and incorporates rationales during inference. We use one additional model \base{} for initial rationale extraction, rationale filtration, and probability estimation of potential next reasoning steps during inference.
% from reasoning datasets and pre-training data. \esti{} also estimates token probabilities for filtering extracted rationales and evaluating candidate reasoning steps. 
These LLMs can be implemented using various state-of-the-art models, allowing for adaptability to specific research needs and computational resources.

% \daniel{To my understanding you don't' fine-tune $M_\text{Esti}$ . So, isn't $M_\text{Esti}$ and M the same? If so, we can drop  $M_\text{Esti}$ ?}

% As we will be using multiple LLMs throughout the process, we define them here: $M$ is the base pre-trained and aligned LLM used for initial rationale extraction from both reasoning datasets and pre-training data. $M_\text{Esti}$ is employed to estimate token probabilities for filtering extracted rationales and evaluating candidate reasoning steps. $M_\text{Ra}$ is the trained rationale model (\mymethod{}) that generates rationales and heuristics during inference. $M_\text{Agent}$ is the general-purpose reasoning agent that produces candidate reasoning steps and incorporates rationales during inference. These LLMs can be implemented using various state-of-the-art models, allowing for adaptability to specific research requirements and computational resources.

% \dongweicomment{it feel strange to define them and explain their roles without diving deep into the process}\daniel{Seems fine to me.}

% First, we explain how rationales are extracted from the unlabeled text (\S\ref{sec:rationale_distillation}), then we proceed to use them to train \mymethod{} (\S\ref{sec:world_model_training}), and finally, we utilize \mymethod{} to supervise the reasoning process at \emph{inference} time (\S\ref{sec:world_model_inference}).

\subsection{Large-scale Rationale Extraction}
% \subsection{Large-scale extraction of rationales}
% \subsection{Rationale extraction and filtration}
% \subsection{Rationale distillation and filtration}
\label{sec:rationale_distillation}

% 1  Training is easy, don't need to rationales on the fly, makes the optimization better
% 2  We can supervise rationales to be humanly readable, which is helpful to xAI
% 3  Goal different, quiet-star is trying to train a model that's better at reasoning, we're trying to train a world model that all LLMs to do reasoning on diverse tasks
% 4  Cost?

% \begin{figure*}[t]
%   \centering
%   % Adjust widths to fit your document
 
%     \centering
%     \includegraphics[trim=1.5cm 1.7cm 1.5cm 2.2cm, width=0.90\linewidth]{figs/GSM_8K_example.pdf} % Second figure file name
%     \caption{An example of the prompt and in-context learning demonstration used to extract rationales (\S\ref{sec:rationale_distillation}) for GSM8K. The rationales in bold represent the implicit thought process during problem-solving. By explicitly extracting these rationales, we're able to train \mymethod{} to predict them given the preceding context.}
%     % \jack{maybe you can add more color to the rationale <BOT> part ot make it look more important}
%     %, which helps in \textbf{better understanding and supervising model's reasoning process.}}
%   % \caption{Two figures side by side}
%   \label{fig:sampling_gsm8k_example}
% \end{figure*}

Implicit rationales are often embedded in unlabelled text, reflecting natural thought processes in daily communication. Our extraction process, illustrated in  \autoref{fig:main_model_step_1}, aims to make these rationales explicit. Using an aligned language model \base{}, we generate rationales from text and then use \base{} to filter these rationales to retain only those that are useful, akin to the self-supervised ``tool'' learning approach described by \citet{schick2023toolformer}. The same \base{} is subsequently used to train \rationalyst{}.

\noindent\textbf{\circled{A} Extracting rationales from pre-training data.}
We employ \base{} to generate rationales from the Pile. Due to the size of this dataset, we implement a pre-filtering process to identify reasoning-rich documents by (1) computing the average semantic embedding of representative reasoning training sets using a paragraph embedding model, and (2) selecting documents from unlabelled datasets that exceed a cosine similarity threshold $\alpha$ when compared to this average embedding.
After pre-filtering, we segment the selected paragraphs into 2000-word segments and instruct \base{} to generate rationales at the end of each sentence, using prompts with demonstrations.
Detailed information on the prompts and in-context learning demonstrations used for rationale extraction can be found in \autoref{app:prompts}.

\noindent\textbf{\circled{B} Extracting rationales from reasoning datasets.}
In parallel to \circled{A}, we also extract rationales from existing reasoning datasets to adapt the extracted rationales to our tested domain and stabilize training. For a given reasoning dataset with pairs of questions and final answers \( D = \{(q_i, a_i)\}_{i=1}^{m} \), we create a prompt \( P \) that instructs 
% a powerful LLM 
\base{} to generate rationales for each reasoning step in the final answer \( a_i \). 
The input of the prompt consists of the entire question and answer, and the output includes implicit rationales that can be inferred from the reasoning process in the answer.
Consider the concrete example from existing datasets (bottom) in \autoref{fig:main_model_step_1}. The solution involves two reasoning steps: ``Natalia sold 48 / 2 = 24 clips in May'' and ``Natalia sold 48 + 24 = 72 chips altogether.'' Here, the implicit rationale that connects the first and second steps, \textit{``Now we should calculate the sum of chips in April and May,''} is implicit yet helpful for the prediction of the second step. These rationales are subsequently filtered and used to train \mymethod{}.

% \noindent\textbf{\circled{B} Extracting rationales from pre-training data.}
% In parallel to \circled{A}, we also use \base{} to generate rationales for web-scale \emph{unlabelled} datasets \citep{gao2020pile} used for pre-training modern LLMs because existing reasoning datasets cover only a small amount of text data. 
% Due to the size of those datasets, we perform pre-filtering to identify documents rich in reasoning by (1) calculating the average semantic embedding of representative reasoning training sets using a paragraph embedding model and (2) selecting documents from unlabelled datasets with cosine similarity above $\alpha$ to this average embedding. Details on the prompts and in-context learning demonstrations we used for rationale extraction are provided in Appendix~\ref{app:prompts}.

% \begin{figure*}[t]
%   \centering
%   % Adjust widths to fit your document
 
%     \centering
%     \includegraphics[trim=4.0cm 18cm 17cm 0.1cm, width=0.93\linewidth]{figs/world model step2_1.pdf} % Second figure file name
%     \caption{Still need to work on it.}
%     \label{fig:main_model_step_2}
% \end{figure*}

\noindent\textbf{\circled{C} Filtering extracted rationales.} Generated rationales in \circled{A} and \circled{B} may not always be accurate or helpful. In reasoning tasks, our objective is for the extracted rationales to effectively aid in future reasoning, which means a good rationale should enhance the likelihood of accurately predicting the following text. 
% Given that these rationales are embedded within the text, effective reasoning in this context means generating the correct subsequent text, 
Let $i$ be the position of the rationale $r$ in the sequence $\mathbf{x} = x_1, \ldots, x_n$. Given a sequence of weights $(w_k)_{k\in\mathbb{N}}$, the cross-entropy loss for future token prediction is defined as:
\[
L_i(\mathbf{r}) = -\sum_{j=i}^n w_{j-i} \cdot \log {p_{M}(x_j \mid \mathbf{r}, x_{1:j-1})},
\]
where \esti{}, in a different role from its previous use, is employed to estimate the probability over tokens $x_i, \ldots, x_n$ prefixed by preceding tokens $x_{1:i-1}$ and rationale $\mathbf{r}$. The weight assigned to each future token decreases exponentially by a factor of 0.9 for each step further away it is from the rationale. We compute $L_i =  L_i(r_i) - L_i(\varepsilon)$, where $\varepsilon$ represents an empty rationale (i.e. predicting following tokens based only on preceding tokens). A rationale is considered helpful if it makes the prediction of future tokens easier, indicated by $L_i \geq \tau_f$, where $\tau_f$ is a filtering threshold. We retain rationales for which adding the rationale \emph{reduces} the loss by at least $\tau_f$ compared to having no rationale.

It's crucial to clarify two key aspects of our rationale extraction process. First, while \base{} extracts rationales from the training sets of reasoning datasets, these training sets are not directly used as targets when training \rationalyst{}. Instead, the target of training is the rationale generated by \base{}. Second, we explicitly instruct \base{} to exclude answers from the extracted rationales. This precaution prevents answer leakage in our prompts.

\subsection{\mymethod{} Training} \label{sec:world_model_training}
The goal of \mymethod{} (denoted by \rationalyst{}) training is to develop a model that can generate implicit rationales to guide stepwise problem-solving during inference time.
For web-scale datasets like The Pile, the input context consists of a segment of text from a document. \rationalyst{} learns to generate an implicit rationale that can guide the prediction of the next segment of text in the document's flow. 
In the case of structured reasoning datasets such as GSM8K or ECQA, the input context includes the question and any preceding reasoning steps toward the answer. Here, \rationalyst{} learns to generate rationales that could guide the next step in the problem-solving sequence.

Given the appropriate context from either source, the implicit rationales, extracted and filtered as described in \S\ref{sec:rationale_distillation}, serve as the target outputs during training.
The overall training objective is to minimize the per-token cross-entropy loss between the generated rationales and their ground truth values from the extracted and filtered rationales.
%By learning to generate appropriate rationales for both free-form text and structured problem-solving data, \mymethod{} develops the ability to provide meaningful guidance across a wide range of contexts during inference.

% To illustrate our experiment, we provide a concrete example of the input and output during training and the heuristics used during inference. \dongweicomment{I've removed the figure}. The input for \rationalyst{} training is: ``Question: Michael had 58 golf balls... How many golf balls did he have at the end of Wednesday? Answer: Michael started with 58 golf balls. After losing 23 on Tuesday, he had 58 - 23 = 35." The corresponding label for \rationalyst{} training is: ``Since Michael now has 35 balls, the next calculation should start from 35, not 58. After losing 2 more on Wednesday, we need to find out how many golf balls he had left." During inference, given the next reasoning step ``After losing 2 more, he had 58 - 2 = 56 golf balls", the probability heuristic for it, estimated by \esti{} conditioned on the implicit rationale generated by \rationalyst{}, is 0.87. The step-based heuristic in this case is -1, indicating there is one remaining step until the final answer.
% Apart from 

\subsection{Inference with the Help of \mymethod{}}\label{sec:world_model_inference}
During inference, any general-purpose LLM (the ``agent model" or \agent{}) can be employed for reasoning across various problems. Algorithm 1 outlines the procedure.

\agent{} generates reasoning incrementally in a chain-of-thought fashion, producing multiple candidates for the next reasoning step. These steps and the question form a ``reasoning trajectory" $T$ that aims to solve the problem, which also serves as input to \rationalyst{}. \rationalyst{} then generates $r$, the implicit rationale (line 3) 
% \yining{Might be helpful the combine this with last sentence}.
With the help of implicit rationale, we provide supervision for the next reasoning step. Two supervision methods we considered are: 
% \yining{Any citations to support the facts that these two supervisions exist? Could from other LLM papers or any CogSci journal.}: % \dongwei{added for explicit supervision}

\paragraph{Implicit supervision.} 
% \yining{If this is the first time introducing the term "implicit supervision," you might want to briefly introduce it and why it's important for rationalists?}
For this supervision, \agent{} generates the next reasoning steps conditioned on the trajectory $T$ (line 6). We then use \base{} to estimate the probability of potential next reasoning steps given rationale $r$ and reasoning trajectory $T$ (line 13). This probability-based heuristic aligns with our rationale filtration process used during \rationalyst{} training: just as we identified rationales that improved the prediction of future text during filtration, here we use rationales to improve the selection of future reasoning steps. By leveraging the probability estimates as a heuristic, we can discriminate between more and less likely next steps in the reasoning process, guiding the overall trajectory towards more accurate conclusions.

\paragraph{Explicit supervision.} 
%\yining{Similar suggestions as implicit supervision}
Another approach is to directly incorporate the implicit rationale into the generation of the next reasoning steps. This method makes the previously implicit rationale an explicit part of the reasoning process. To do that, we ask \agent{} to generate multiple candidate next steps by temporarily appending $r$ to the trajectory $T$, and then producing potential continuations based on this augmented context (line 8). Then, we estimate the probability of candidate generations according to \agent{} (line 15). This approach allows \agent{} to make the final decision on the next reasoning step, as in normal beam search \citep{snell2024scalingllmtesttimecompute, yao2023tree}, while benefiting from the additional context provided by \rationalyst{}'s rationales.

\algrenewcommand\algorithmicrequire{\textbf{}}
\algrenewcommand\algorithmicensure{\textbf{}}

\renewcommand{\algorithmiccomment}[1]{\hfill$\triangleright${\color{gray}{\textit{#1}}}}

\newcommand{\MAgent}{\textcolor{mina-blue}{M_\text{Agent}}}
\newcommand{\MRa}{\textcolor{mina-yellow}{M_\text{Ra}}}
\newcommand{\M}{\textcolor{mina-red}{M}}

\makeatletter
\renewcommand{\ALG@beginalgorithmic}{\tiny}
\makeatother

\makeatletter
\algnewcommand{\LineComment}[1]{\Statex \hskip\ALG@thistlm \(\triangleright\) #1}
\makeatother

\algrenewcommand\alglinenumber[1]{\small #1:}

% \begin{figure}[h]
% \centering
\noindent
\resizebox{\linewidth}{!}{
\begin{minipage}{1.13\linewidth}
\begin{algorithm}[H]
\captionsetup{font=normalsize} % Apply small font to caption
    \caption{\small{Inference with \mymethod}}
    \label{alg:cond_gen}
    \begin{algorithmic}[1]
    \Statex \small{\textbf{Input:} Question $q$, \mymethod{} \textcolor{mina-yellow}{$M_\text{Ra}$}, Agent model \textcolor{mina-blue}{$M_\text{Agent}$}, Probability estimation model \textcolor{mina-red}{$M$}};  
    \Statex \small{\textbf{Functions:} Heuristic function $H(\textcolor{mina-yellow}{M_\text{Ra}}, q, T)$, stopping condition  $stop\_condition()$}
    \Require \small\textbf{Hyperparameters:} Sampling temperature $t$ and number of sampled rationales $N$
    \State $T \gets q$  
    \Comment{Initialize reasoning trajectory as the question.}
    \Repeat
        \State $r \gets \MRa(T)$  \Comment{\parbox[t]{.62\linewidth}{Generate rationale given trajectory.}}
        \State $heuristic\_list = \emptyset $
        \If{$supervision==implicit$}
            \State $next\_steps \gets \MAgent(T)$ 
        \ElsIf{$supervision==explicit$}
            \State $next\_steps \gets \MAgent(T, r)$ 
        \EndIf
        \For{$n = 1 \ldots N$}
            \State $x \gets next\_steps[n]$  
            \Comment{Take next step generation.}
            \If{$supervision==implicit$}
                \State $h \gets \M(x | T, r)$ \Comment{\parbox[t]{.43\linewidth}{Estimate probability of next reasoning step.}}
            \ElsIf{$supervision==explicit$}
                \State $h \gets \MAgent(x | T, r)$ 
            \EndIf
            \State $heuristic\_list.append(h) $ 

        \EndFor
        
            % \State $r \gets H(M_\text{Ra}, q, T + x)$ \Comment{\parbox[t]{.45\linewidth}{Get the heuristic with extended reasoning trajectory.}}

            % \small{\color{gray}{\hspace{0.18cm} 
                        % \# Get heuristic with extended reasoning trajectory}}
        \State $m\_idx \gets argmax(heuristic\_list) $
    \State $T \gets T + next\_steps[m\_idx]$
        \Comment{\parbox[t]{.38\linewidth}{
            Extend trajectory with the highest scoring step.
        }}
    \Until { $stop\_condition(T)$}
            \Comment{\parbox[t]{.52\linewidth}{
                E.g., the trajectory contains strings like ``The final answer is:''
            }}
            
    \State \textbf{return} $T$
    \end{algorithmic}
\end{algorithm}
\end{minipage}
}
% \end{figure}

% \yining{In the algorithm, you may not want the indent for Input, Function, and Hyperparameters. Try \textbackslash begin{flushleft} command}

After providing heuristics for the next reasoning steps, the step with the highest heuristic (line 19) is selected. The reasoning trajectory is then extended with this highest-scoring step (line 20). The reasoning process concludes when the stop condition is satisfied (line 21), which varies by dataset and often includes cues like ``The final answer is:" that can be specified in system prompts for \agent{} for different tasks.

The computational cost of \rationalyst{} is comparable to a normal beam search, with the only additional cost being the generation of rationales, which are typically quite short.

\section{Experimental Setup}

% \subsection{Experimental Setup}
\subsection{Setup for Training \mymethod{}}

% The summary of our experiment setup can be found in \autoref{tab:experimental_setup}. We will expand more on each part below.

% \begin{table*}[h]
%     \centering
%     \begin{tabular}{lccc}
%         \toprule
%         \textbf{Dataset}  &\textbf{\# Total rationales} &\textbf{Filter \%} & \textbf{Rationale Precision}  \\
%         \midrule
%         % GSM8K        & 23388    & 26.3\% &  95.2\% \\
%         GSM8K  & 17566  & 19.5\% & 95.2\% \\
%         % filtered: 5156, total: 22011 0.23424651310708283
%         % APPS & &   \\
%         ECQA  & 19669 &  57.6\%  & 95.1\% \\
%         % OpenWebMath &  & \\
%         \bottomrule
%     \end{tabular}
%     \caption{Total number of generated rationales, the percentage of filtered rationales, and the precision we control for filtration for each dataset. Precision is calculated separately from 100 human annotations on each dataset}
%     \label{tab:experimental_setup}
% \end{table*}

\paragraph{Rationale extraction.} 
As discussed in \S\ref{sec:rationale_distillation} \circled{A}, we perform pre-filtering on The Pile, an unlabelled web-scale dataset, to identify documents with extensive reasoning content before rationale extraction. This is achieved by computing the average semantic embedding from the training sets of the reasoning datasets we test, filtering documents that exceed the cosine similarity threshold $\alpha$ of 0.3, and keeping only the documents with length under 2000 tokens to fit within LLaMa-3 models' context length. The model we used to calculate these embeddings is {\tt MPNet-base} \citep{song2020mpnet}.

Following the recipe in \S\ref{sec:rationale_distillation} \circled{B}, we also extract rationales from existing reasoning datasets. GSM8K \citep{gsm8k} and ECQA \citep{aggarwaletal2021ecqa} were selected for their complementary coverage of mathematical and commonsense reasoning, respectively. This combination ensures \mymethod{} is trained on diverse reasoning patterns, enhancing its versatility across various tasks.

% \daniel{incorporate these: }
% \circled{A}
% \circled{B}

\paragraph{Rationale annotation and filtration.} The model \base{} used for rationale extraction and rationale filtering are both LLama-3-8B-Instruct \cite{llama3}. On GSM8K and ECQA, we manually annotated 100 pairs of \{\text{preceding}\_\text{context}, rationale, \text{following}\_\text{context}\} to determine an appropriate filtration threshold. The annotations include 50 positive and 50 negative rationale examples. Since it's straightforward to scale up the extraction of rationales from unlabelled data for filtration, we prioritize maximizing the precision of our filtered rationales, even if it means extracting fewer of them. We set the threshold \(\tau_f\) to ensure that 95\% of the filtered rationales are accurate. On The Pile, we do not perform rationale annotation due to its diverse composition of corpora with varying characteristics. So the filter threshold \(\tau_f\) for the Pile is set to 0 for all of its subdomains. The results of rationale extraction and filtration on GSM8K, ECQA, and The Pile are presented in \autoref{tab:filter_result}. From our extraction and filtration process, we obtained approximately 14k rationales from GSM8K and ECQA combined, and about 65k from The Pile after filtration. Detailed statistics of rationale extraction and filtration results for each dataset, including the number of rationales per document and filtration rates, can be found in \autoref{resulting_data}.

{
\setlength{\tabcolsep}{1.4pt}
\begin{table}[ht]
\small
    \centering
    \resizebox{\linewidth}{!}{
    \begin{tabular}{lccccc}
        \toprule
        \textbf{Dataset} &\textbf{Subdomains} & \textbf{\# Docs.} &\textbf{\# Rationales} &\begin{tabular}{@{}c@{}} \textbf{Rationales} \\ \textbf{Left (\%)}\end{tabular}   & \textbf{$\tau_f$}  \\
        \midrule
        % GSM8K        & 23388    & 26.3\% &  95.2\% \\
        GSM8K  & N/A & 7473 & 17566  & 19.5 & 1.2 \\
        % filtered: 5156, total: 22011 0.23424651310708283
        % APPS &N/A & 6978 & 42.7\% & 0.6 \\
        ECQA  & N/A  & 7600 & 19669 &  57.6  & 0.5 \\
        \midrule
        \multirow{6}{*}{The Pile}&  Pile-CC & 266.6K & 853.2K & 2.9 & 0\\
        & StackExchange& 21.8K& 113.6K& 29.8 & 0\\
        & Github & 19.9K & 45.8K & 2.6& 0\\
        & HackerNews& 5.8K&  24.4K& 9.4& 0\\
        & PubMed Central & 4.9K & 18.6K & 3.2 & 0\\
        & Wikipedia (en)& 4.2K& 23.0K & 7.8 & 0\\
        % & PubMed Abstracts & 3257 & \\
        % & Enron Emails & 2637 & \\
        % & USPTO Backgrounds & 620 & \\
        % & Ubuntu IRC& 48&  & \\
        \bottomrule
    \end{tabular}
    }
    \caption{\textbf{The statistics on rationale sampling and filtration.} We provide the total number of documents and rationales before filtering, and the percentage of leftover rationales after filtering.}
    \label{tab:filter_result}
\end{table}
}

\paragraph{\mymethod{} training.} \mymethod{} is fine-tuned with LLaMa-3-8B-Instruct  as the base model. We use the default hyperparameters as specified in the LLaMa-3 technical report \citep{llama3} for fine-tuning. After training, we conducted manual annotation of the model's output and found that the accuracy of rationales generated on unseen test data closely matches the filtration accuracy we specified for training data through our filtration parameters.

\subsection{Setup for Evaluating \mymethod{}}
\label{subsec:evaluations:in:experiments}

We evaluate \mymethod{} across diverse reasoning tasks including mathematical (GSM8K \citep{gsm8k}, MATH \citep{math}, commonsense (ECQA\citep{aggarwaletal2021ecqa} and HellaSwag \citep{zellers2019hellaswag}), logical (ProofWriter \citep{proofwriter}), scientific (ARC \citep{clark2018think}), and multi-task reasoning (MMLU-Pro \citep{wang2024mmlupro}). For inference, we use LLaMa-3-8B-Instruct as our base agent model with temperature 0.7 and top-k sampling of 3. We compare \mymethod{} against both process supervision approaches (using LLaMa-3-8B-Instruct and GPT-4) and outcome-based verifiers. Complete experimental details are provided in \autoref{appendix:eval_setup}.

\subsection{Setup for other verifiers.} 
To evaluate the effectiveness of \mymethod{}'s process supervision, we compare it with other approaches. For \emph{process supervision} with other models, we include LLaMa-3-8B-Instruct and GPT-4 in our comparison. These models are prompted to rerank partial reasoning trajectories as reasoning steps are generated. The prompts and in-context learning demonstrations used for these models on representative datasets are provided in \autoref{app:llama3-reranking}. For \emph{outcome supervision}, we also compare with outcome-based verifiers derived from LLaMa-3-8B-Instruct. These verifiers are fine-tuned on the training sets of each reasoning dataset. Following the approach outlined by \citet{gsm8k}, they assess the correctness of the final prediction by directly evaluating the question and final solution.
This comparison allows us to assess the performance of \mymethod{} against both process-based and outcome-based supervision methods.
\section{Empirical Results}
\label{section:results}

\subsection{Main result: \mymethod{} Improves Performance on Various Tasks} 

% \subsection{Training on web-scale unlabeled data enhances reasoning} 

\label{sec:big_world_model}
% check if it will generalize to new datsets\\
% add greedy reuslt to agent candidates
%\dongweicomment{also test zero-shot cot} 
In this section, we train \mymethod{} using a combination of rationales extracted from GSM8K and ECQA, as well as from The Pile, as outlined in \autoref{tab:filter_result}. The baseline does not use any verifier. We use implicit supervision for this experiment. The main result is shown in \autoref{tab:big_world_model}.

% \begin{table*}[h]
% \small
%     \centering
%     \begin{tabular}{llccccc}
%         \toprule
%         \multirow{2}{*}{\textbf{Reasoning Type}} &  \multirow{2}{*}{\textbf{Dataset}} &  \multirow{2}{*}{\textbf{Baseline}} & \multicolumn{2}{c}{\textbf{Model - Mixed}} & \multicolumn{2}{c}{\textbf{Model - All}} \\
%         & & & \textbf{Acc.} & \textbf{Rel. $\uparrow$} & \textbf{Acc.} & \textbf{Rel. $\uparrow$}\\
%         \midrule
%         Mathematical & GSM8K        & 72.3 & 75.2 & 4.0\%  &\textbf{76.2} & \textbf{5.4\%}\\
%                       & Math        & 28.0 & 31.8 & 13.6\% &\textbf{32.5} & \textbf{16.1\%}\\
%         CommonSense   & ECQA        & 72.6 & 75.5 & 4.0\%  &\textbf{76.2} & \textbf{5.0\%}\\
%                       & HellaSwag   & 58.2 & 59.1 & 1.5\%  &\textbf{59.4} & \textbf{2.1\%}\\
%         Logical       & ProofWriter & 86.4 & 88.2 & 2.1\%  &\textbf{90.1} & \textbf{4.3\%}\\
%         Scientific    & ARC         & 76.6 & 78.8 & 2.9\%  &\textbf{79.3} & \textbf{3.5\%}\\
%         Combined      & MMLU-Pro    & 38.2 & 39.4 & 3.1\%  &\textbf{42.1} & \textbf{10.2\%}\\
%         \bottomrule
%     \end{tabular}
%     \caption{Accuray and relative (rel.) improvement over base using \mymethod{} - mixed (trained with rationales from GSM8K and ECQA) and \mymethod{} - all (trained with rationales from GSM8K, ECQA, and parts of The Pile as detailed in \autoref{tab:filter_result}). \textbf{\mymethod{} generalizes across different reasoning tasks, showing improved performance with training on unlabelled web-scale data}}
%     \label{tab:big_world_model}
% \end{table*}

{
\setlength{\tabcolsep}{4.2pt}
\begin{table}[h]
\small
    \centering
    \resizebox{\linewidth}{!}{
    \begin{tabular}{llccc}
        \toprule
        \multirow{2}{*}{\textbf{Reasoning Type}} &  \multirow{2}{*}{\textbf{Dataset}} &  \textbf{Baseline}  & \multicolumn{2}{c}{\textbf{\mymethod{}}} \\
        & & \textbf{Acc.} &  \textbf{Acc.} & $\Delta$ \textbf{Acc.}\\
        \midrule
        Mathematical & GSM8K        & 77.6 &\textbf{81.6} & \textbf{4.0}\\
                      & Math        & 28.0 &\textbf{32.5} & \textbf{4.5}\\
        CommonSense   & ECQA        & 72.6 &\textbf{75.2} & \textbf{2.6}\\
                      & HellaSwag   & 58.2 &\textbf{60.3} & \textbf{2.1}\\
        Logical       & ProofWriter & 86.4 &\textbf{90.7} & \textbf{4.3}\\
        Scientific    & ARC         & 77.6 &\textbf{80.7} & \textbf{3.1}\\
        Combined      & MMLU-Pro    & 39.6 &\textbf{45.3} & \textbf{5.7}\\
        \bottomrule
    \end{tabular}
    }
    \caption{Accuracy and absolute improvement over baseline using \mymethod{}. 
    % (trained with rationales from GSM8K, ECQA, and parts of The Pile as detailed in Table \ref{tab:filter_result}). 
    \textbf{\mymethod{} generalizes improved performance across different reasoning tasks.}}
    \label{tab:big_world_model}
\end{table}
}

% \paragraph{\mymethod{} generalizes to various reasoning tasks.}
Evaluation of \mymethod{} shows that training with rationales from GSM8K, ECQA, and The Pile improves performance not only on GSM8K and ECQA, but also on other reasoning tasks (e.g. scientific reasoning, logical reasoning, etc) not directly used in rationale extraction. This supports the idea that rationales can be broadly applicable across different reasoning tasks. In addition, since we use the same model (LLaMa-3-8B-Instruct) for rationale extraction, filtering, \mymethod{} training, and inference, our results do not leverage external knowledge from stronger models like LLaMa-3-70B-Instruct or GPT-4. Future work might change \base{} to stronger models, with the expectation that higher-quality rationales will lead to better performance.

\subsection{Ablation: Web-scale Rationales Enhance Performance Across Tasks} \label{sec:ablation_web_scale_rationales}

To assess the benefit of web-scale rationales, we train another model: \mymethod{} w/o Pile solely on rationales extracted from the training sets of GSM8K and ECQA. We re-ran the experiments on the same reasoning datasets using implicit supervision. The results are detailed in \autoref{tab:ablation_pile}.
{
\setlength{\tabcolsep}{4.5pt}
\begin{table}[h]
\small
    \centering
    \begin{tabular}{lcC{1.9cm}c}
        \toprule
        \multirow{2}{*}{\textbf{Dataset}} & \multirow{2}{*}{\textbf{\mymethod{}}}  & \textbf{\mymethod{} (w/o Pile)} & \multirow{2}{*}{$\Delta$\textbf{Acc.}} \\
        % & & \textbf{Acc.} & $\Delta$\textbf{Acc. $\downarrow$}\\
        \midrule
         GSM8K       & \textbf{81.6}  &80.3 & \textbf{-1.3}\\
         Math        & \textbf{32.5}  &31.4 & \textbf{-1.1}\\
         ECQA        & \textbf{75.2}  &74.5 & \textbf{-0.7}\\
         HellaSwag   & \textbf{60.3}  &59.1 & \textbf{-1.2}\\
         ProofWriter & \textbf{90.7}  &88.2 & \textbf{-2.5}\\
         ARC         & \textbf{80.7}  &78.8 & \textbf{-1.9}\\
         MMLU-Pro    & \textbf{45.3}  &41.2 & \textbf{-4.1}\\
        \bottomrule
    \end{tabular}
    \caption{
    An ablation study on the benefit of rationales extracted from pre-training data (The Pile).
    The consistent accuracy drop shows that, \textbf{utilizing web-scale rationales improves performance on various reasoning datasets.}}
    \label{tab:ablation_pile}
\end{table}
}

We find that training the model on web-scale data results in better performance compared to training only on the rationales extracted from GSM8K and ECQA. This improvement is consistent and particularly significant on MMLU-Pro.
Web-scale data likely provides exposure to more diverse reasoning types and content, including specialized knowledge, complex real-world scenarios, and interdisciplinary connections not present in the more focused datasets.

\subsection{Ablation: Implicit Supervision Works Better than Explicit Supervision} \label{sec:performance_boost}
In this section, we conduct ablation studies to test the effectiveness of different supervision methods. To isolate the impact of supervision methods and minimize confounding variables, we focus on GSM8K and ECQA as representative benchmarks for mathematical and commonsense reasoning, respectively. 
We train two versions of  \mymethod{}: one on rationales extracted from the GSM8K training set (\mymethod{} - GSM8K) and another on rationales from the ECQA training set (\mymethod{} - ECQA). These models are used to supervise \agent{} during inference on their respective tasks.

As shown in \autoref{tab:ablation_on_heuristic_choice}, implicit supervision outperforms explicit supervision. Our manual analysis revealed that implicit supervision's superior performance stems from its greater robustness to errors. When \mymethod{} generates an imperfect rationale, the probability-based heuristic used in implicit supervision can still provide useful guidance even if the rationale itself is not ideal. This approach is less likely to lead \agent{} to produce incorrect next steps. In contrast, explicit supervision directly incorporates potentially flawed rationales into the reasoning process, which can cause \agent{} to produce incorrect next steps. Essentially, implicit supervision acts as a softer guide, allowing for some imperfection in rationales, while explicit supervision more strictly adheres to potentially flawed rationales, making it more susceptible to errors.

\begin{table}[h!]
\small
    \centering
    \begin{tabular}{lcc}
        \toprule
         \textbf{Heuristic}$\downarrow$ \; - \;  \textbf{Evaluation task}$\rightarrow$  & \textbf{GSM8K} & \textbf{ECQA} \\
        % \midrule % Only spans columns 2-5, avoiding intersecting the multirow
        \midrule % Only spans columns 2-5, avoiding intersecting the multirow 
        Implicit Supervision & \textbf{80.3} & \textbf{74.5}\\
        Explicit Supervision & 77.5 & 72.2 \\
        \bottomrule
    \end{tabular}
    \caption{Comparison of implicit and explicit supervision methods on GSM8K and ECQA tasks. \textbf{Implicit supervision outperforms explicit supervision due to its robustness to errors.}}
    
    \label{tab:ablation_on_heuristic_choice}
\end{table}

\subsection{\mymethod{} Outperforms Other Verifiers}\label{sec:comparsion_with_verfiers}
\autoref{tab:comparsion_verifier} presents an analysis of \mymethod{} against various verifiers. Our findings reveal several insights:

\paragraph{\mymethod{} outperforms vanilla LLaMa-3-8B-Instruct using process supervision:} \mymethod{}, even without leveraging The Pile dataset, outperforms process-based verifiers using vanilla LLaMa-3-8B-Instruct. A manual examination of reasoning trajectories suggests that LLaMa-3-8B-Instruct faces difficulties in reranking partial reasoning steps. This challenge likely stems from the model's struggle to differentiate among its own generated outputs, a phenomenon observed in recent studies \citep{jiang2024selfincorrect, cant}.

\begin{table}[ht]
\small
    \centering
        \resizebox{1.00\linewidth}{!}{
    \begin{tabular}{l@{\hspace{0.4em}}c@{\hspace{0.4em}}c}
        \toprule
        \textbf{Supervision} & \textbf{GSM8K} & \textbf{ECQA} \\
        \midrule
        N/A & 77.6 & 72.6\\
        Process Supervision w/ LLaMa-3 & 77.4 & 71.5\\
        Process Supervision w/ GPT-4 & 80.0 & 74.7 \\
        Outcome Supervision w/ LLaMa-3 + FT & 79.2 & 74.3\\
        \mymethod{} w/o Pile & 80.3 & 74.5\\
        \mymethod{} & \bf{81.6} & \bf{76.2} \\
        \bottomrule
    \end{tabular}}
    \caption{Comparison of different supervision methods. \textbf{\mymethod{} outperforms both strong verifiers like GPT-4 and similarly-sized models fine-tuned on matching training data.}
    }
    \label{tab:comparsion_verifier}
\end{table}

\paragraph{\mymethod{} shows superior process-supervision performance than much bigger models like GPT-4:} 
From \autoref{tab:comparsion_verifier} we 
observe consistent superior performance of \mymethod{} compared to GPT-4's process supervision. We hypothesize that this advantage arises from \mymethod{}'s specialized design for providing supervision, in contrast to GPT-4's general-purpose training.

\paragraph{\mymethod{} surpasses outcome-based verifiers trained using matching data:} Notably, our method surpasses the performance of fine-tuned outcome-based verifiers on both GSM8K and ECQA datasets, despite these verifiers being trained on matching data. We attribute this success to the richer feedback provided by process-based supervision compared to outcome-based approaches.

We would also like to mention that methods like self-consistency and other prompting techniques (e.g. STEP-BACK Prompting \citep{zheng2024stepbackevokingreasoning}) are complementary to our approach. \mymethod{} can be integrated with these methods to potentially achieve even better results. For more detail, please refer to \autoref{self-consistency}.

\subsection{\mymethod{} Generates Accurate and Easy-to-understand Rationals} \label{sec:intepreability}
We annotate some samples from the test set of Math \citep{math} at inference time, which was not part of the rationale sampling datasets. Through manual observation, we find that our model can generate useful rationales that is helpful for understanding LLM's reasoning process on Math (an example is provided in \autoref{app:example_rationale}). Comparing the rationales generated by \mymethod{} with those generated by Quiet-Star \citep{zelikman2024quietstar} on the same problems, we find that our method produces more human-understandable rationales. We believe this happens because Quiet-Star optimizes rationales during training using the accuracy of the final prediction as a reward. This approach, while effective for improving task performance, does not explicitly prioritize human interpretability. In addition, this appraoch might inadvertently develop shortcuts or non-intuitive patterns that optimize for accuracy but not necessarily for clarity or human understanding.
% We suspect this is because Quiet-Star utilizes the final prediction accuracy as a reward for its optimization which may inadvertently lead to shortcuts (i.e., reward hacking) during its training
% and does not always ensure they are easily understandable by humans.
% on optimizing rationales based on final prediction accuracy during training, which may utilize shortcuts 

\subsection{\mymethod{} Introduces Minimal Inference Cost Overhead}
\label{sec:compute_analysis}
We analyzed the inference costs of \mymethod{} compared to baseline approaches. Using vLLM for efficient deployment with LLaMA-3-8B-Instruct as both the agent model and \mymethod{}, we find minimal time overhead (within 5\%) despite requiring two model calls per reasoning step, achieved through async batched execution and short rationale generation. Memory requirements approximately double with LLaMA-3-8B-Instruct, but this overhead becomes proportionally smaller with larger agent models (e.g., when using LLaMA-3-70B, \mymethod{}'s additional 8B parameters represent a much smaller relative increase). These computational costs are minimal compared to \mymethod{}'s demonstrated accuracy improvements, making the method efficient for applications prioritizing reasoning quality.
\section{Discussion} \label{app:discussion}
% \paragraph{Can models already improve their reasoning by training on additional data without the help of \mymethod?}

% In addition, even pretraining on high quality reasoning data is unlikely to surpass our self-supervised approach, the whole focus of which is to supervise reasoning during training.
% We highlight the results in \autoref{tab:big_world_model}, where finetuning on our rationales improves performance overall compared to LLaMa-3. 

% We are skeptical about this. While training on extra data that includes reasoning patterns has been shown to enhance models' reasoning capabilities, such data are usually already part of the models' training sets. Additionally, an autoregressive approach may not necessarily enhance reasoning skills. There is also a "generality tax," which refers to the model's performance penalty for being more general-purpose (which will happen when it's trained on general data). In contrast, our method is tailored specifically to supervise reasoning.

% High quality reasoning data is difficult to collect. Since our approach is self-supervised, we can create rich, diverse training data on the fly, which removes any bottlenecks caused by limited data availability.

\paragraph{Scaling up \mymethod{}.} Scaling \mymethod{} with stronger models and increased computational resources is a logical next step. Utilizing stronger models, such as LLaMa-3-70B or GPT-4, would enhance the quality of extracted rationales, improve filtration accuracy, and ultimately strengthen \mymethod{}. However, due to computational constraints, we have not pursued this.
% , which remains a limitation of this paper.
Additionally, using larger unlabelled datasets with more extensive reasoning content, such as OpenWebMath \citep{paster2023openwebmath}, is currently infeasible due to the significant computational and time requirements for pre-filtering and training. 
% These enhancements are planned for future work.

\paragraph{Connection to research on scaling test-time compute.}
Recent research has focused on extending computational resources at test-time \citep{snell2024scalingllmtesttimecompute, wu2024empiricalanalysiscomputeoptimalinference}, particularly for complex reasoning tasks. 
In our experiments, we focus on developing heuristics and employ a straightforward approach of sampling multiple candidates and reranking them based on \mymethod{}'s guidance. However, \mymethod{}'s framework is compatible with more sophisticated test-time compute techniques. Its heuristics can be integrated into existing algorithms like beam-search or look-ahead search.
% potentially enhancing their performance without significantly increasing computational cost.

% \mymethod{}'s approach to extracting and utilizing implicit rationales from unlabeled text data also aligns with recent trends in leveraging synthetic data and verifiers for improving LLM reasoning \citep{hosseini2024vstartrainingverifiersselftaught}. Unlike methods that require extensive human annotation, \mymethod{} extracts rationales from unlabeled data, offering a more scalable solution. The extracted rationales could also enhance verifiers with interpretable insights into reasoning processes, beyond binary judgments. By integrating \mymethod{}'s approach, we could potentially improve the quality of synthetic data generation and create more effective self-improvement loops for LLMs.

\paragraph{Is training on extracted rationales necessary?}
% In our approach, we first select a subset of unlabelled data that contains strong reasoning signals, then extract implicit rationales from this data for model fine-tuning.
While it has been demonstrated that training on data with robust reasoning signals can enhance reasoning capabilities on its own \citep{gunasekar2023textbooks, jiang2024leanreasoner}, we believe our method offers additional performance benefits for two reasons. First, many language models have already been trained on datasets like The Pile. The value of fine-tuning on previously encountered text is likely lower than the value of fine-tuning on newly incorporated rationales. Second, implicit rationales encapsulate the reasoning process. Pre-training on these rationales enhances reasoning more effectively than focusing on the whole document.
\section{Conclusion and Future Work}

% In this paper, we introduced \mymethod{}, a novel self-supervised model designed to enhance the reasoning capabilities of LLMs by leveraging hidden rationales extracted from unlabeled text. Our approach centers on the effective extraction and utilization of implicit rationales--those underlying thought processes that are not explicitly stated in the text but can be inferred. By capturing these rationales, \mymethod{} provides a mechanism for process supervision during reasoning, enabling LLMs to reason better.

In this paper, we introduced \mymethod{}, a novel self-supervised model designed to enhance the reasoning capabilities of LLMs by leveraging hidden rationales extracted from unlabeled text. While our results demonstrate the effectiveness of this approach, there are several promising directions for future work, including scaling to larger models and datasets, integrating with test-time compute optimization techniques, and exploring the role of implicit rationales in model training. 
% Through this work, we have demonstrated that extracting and utilizing implicit rationales from unlabeled text can meaningfully improve LLMs' reasoning abilities.
We hope these insights will inspire further research into leveraging latent knowledge in pre-training data to enhance model capabilities.

% We would also like to compare our method with other salient work on the verification of reasoning. 
\section*{Acknowledgments}
This work is supported by ONR grant (N00014-241-2089). The GPUs were provided by the DSAI cluster. 
We sincerely thank Eric Zelikman, Tianmin Shu, and the broader JHU CLSP community for discussions and inspiration.

\section{Limitations}
% \paragraph{Limitations in experimental scope.} 
One limitation of this work is the comprehensiveness of our experiments. In future research, we plan to extend our experiments to a broader range of reasoning tasks and compare \mymethod{} with other outcome-based and process-based verifiers. We also plan to adjust the combination of rationales used to train \mymethod{} by (1) sampling from different reasoning tasks and (2) altering the mix of rationales in unlabelled web-scale pre-training data to better understand its generalizability.  An additional limitation is that we could potentially improve performance through preference tuning (e.g., DPO) where the model learns to distinguish between valid and invalid rationales, but we have not explored this direction.

% \daniel{Somewhere (here or elsewhere) let's also highlight that 
% tools like RORA https://aclanthology.org/2024.acl-long.60/ can be used of annotating or filtering the augmented data here. 
% }

% \newpage
\bibliography{ref}

\newpage

\appendix

\clearpage

\section{Evaluation Setup Details}\label{appendix:eval_setup}
\subsection{Task Configuration and Metrics}
We evaluate \mymethod{} on a diverse set of reasoning tasks, carefully selected to assess performance across different reasoning domains. \autoref{tab:setup_appendix} provides a comprehensive overview of our evaluation configuration.

\begin{table}[h]
    \centering
    \small
    \begin{tabular}{lcccccc}
        \toprule
        %\cmidrule(r){2-4} \cmidrule(r){5-7}
        \textbf{Task}&  \textbf{\#Eval} & \textbf{\#Shots} &   \textbf{Reasoning Type} \\
        \midrule
        % \multirow{3}{*}{GPT-4} 
        \textbf{GSM8K}  & 1319 & 8  & Math    \\
        \textbf{Math}  & 5000 & 5  & Math     \\
        \textbf{ECQA}  & 17944 & 6  & CommonSense   \\
        \textbf{HellaSwag}  & 10000 & 4  & CommonSense    \\
        % \textbf{WinoGrande} & Test & 1319 & 2  & CommonSense Reasoning & \\
        \textbf{ProofWriter}  & 600 & 2  & Logical   \\
        \textbf{ARC}  & 1172 & 4  &  Scientific    \\
        \textbf{MMLU-Pro} & 12000 & 5  &  Mixed  & \\
        % \textbf{APPS} & Test & 700 & 3 & Code Generation  & pass@1 \\
        \bottomrule
    \end{tabular}
    \caption{\textbf{Detailed configuration of evaluation tasks.} For each dataset, we report the training set size (\#Train), evaluation set size (\#Eval), and number of few-shot demonstrations (\#Shots) used during evaluation. We use exact match accuracy across all tasks, following standard evaluation protocols.}
\label{tab:setup_appendix}
\end{table}

For each dataset, we implement specific evaluation protocols:
\begin{itemize}
\item \textbf{GSM8K}: We evaluate on the standard test split, using 8-shot demonstrations that showcase step-by-step mathematical reasoning.
\item \textbf{MATH}: Our evaluation encompasses all difficulty levels and mathematical topics in the dataset, utilizing 5-shot demonstrations calibrated for complexity.
\item \textbf{ECQA}: We use the validation split due to evaluation server constraints for the test set, maintaining consistency with prior work.
\item \textbf{HellaSwag}: Evaluation follows the original setup with 4-shot demonstrations focusing on commonsense inference.
\item \textbf{ProofWriter}: We specifically evaluate on proofs with depth more than 5 to assess complex logical reasoning capabilities.
\item \textbf{ARC}: Our evaluation focuses on the more challenging ARC-Challenge subset, employing 4-shot demonstrations.
\item \textbf{MMLU-Pro}: Following the original paper's protocol, we evaluate across all reasoning categories using 5-shot demonstrations.
\end{itemize}
\subsection{Inference Configuration}
Our inference setup utilizes LLaMa-3-8B-Instruct as the base agent model (\agent{}). Key configuration parameters include:
\paragraph{Sampling Parameters.} We employ temperature = 0.7 and top-k = 3 during inference. This configuration allows for diverse reasoning path exploration while maintaining coherence. The temperature setting represents a calculated trade-off between deterministic output (temperature = 0) and exploration, showing particular effectiveness on logical reasoning tasks like ProofWriter.
\paragraph{Chain-of-Thought Implementation.} For each task category, we implement specialized chain-of-thought prompting:
\begin{itemize}
\item \textbf{Mathematical and Procedural Tasks}: Demonstrations break down solutions into atomic reasoning steps, with explicit intermediate calculations.
\item \textbf{Multiple-Choice Tasks}: Prompts guide systematic analysis of each answer choice, encouraging comparative reasoning.
\item \textbf{Logical Reasoning}: Step-by-step deductive reasoning chains are demonstrated, emphasizing premise utilization.
\end{itemize}
% \subsection{Baseline and Comparison Methods}
% We implement two categories of comparison methods to evaluate \mymethod{}:
% \paragraph{Process Supervision Baselines.} We compare against:
% \begin{itemize}
% \item LLaMa-3-8B-Instruct: Configured identically to our base model but without \mymethod{}'s supervision
% \item GPT-4: Implemented as a high-capacity verifier for reasoning trajectories
% \end{itemize}
% Both models are prompted to rerank partial reasoning trajectories during step generation. Detailed prompting templates and demonstrations are provided in \autoref{app:llama3-reranking}.
% \paragraph{Outcome Verification Baselines.} We develop outcome-based verifiers by:
% \begin{itemize}
% \item Fine-tuning LLaMa-3-8B-Instruct on each task's training set
% \item Implementing direct evaluation of question-answer pairs following \citet{gsm8k}
% \item Maintaining consistent evaluation metrics across all comparison methods
% \end{itemize}
% This comprehensive evaluation framework enables robust comparison of \mymethod{}'s effectiveness against both process-based and outcome-based supervision approaches.

\section{Prompts used for rationale sampling}\label{app:prompts}
\begin{figure*}[ht]
  \centering
  % Adjust widths to fit your document
 
    \centering
    \includegraphics[trim=0.2cm 0cm 0.2cm 0cm, width=\linewidth]{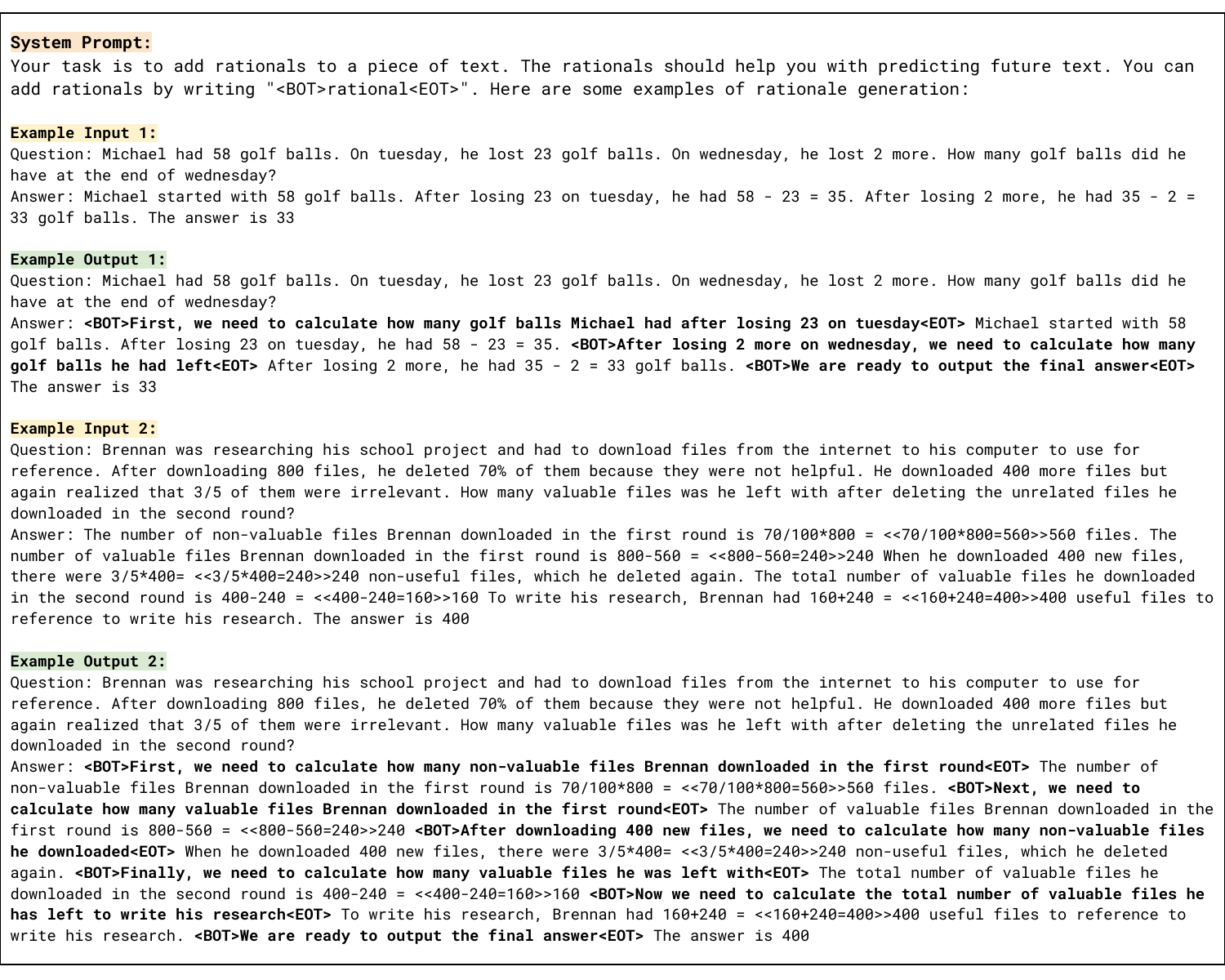} % Second figure file name
    \caption{The prompt and in-context learning examples used for sampling rationales for GSM8K. The bolded rationales represent implicit rationales in the document.}
  % \caption{Two figures side by side}
  \label{fig:sampling_gsm8k}
\end{figure*}

\begin{figure*}[ht]
  \centering
  % Adjust widths to fit your document
 
    \centering
    \includegraphics[trim=0.2cm 1.3cm 0.2cm 1.3cm, width=\linewidth]{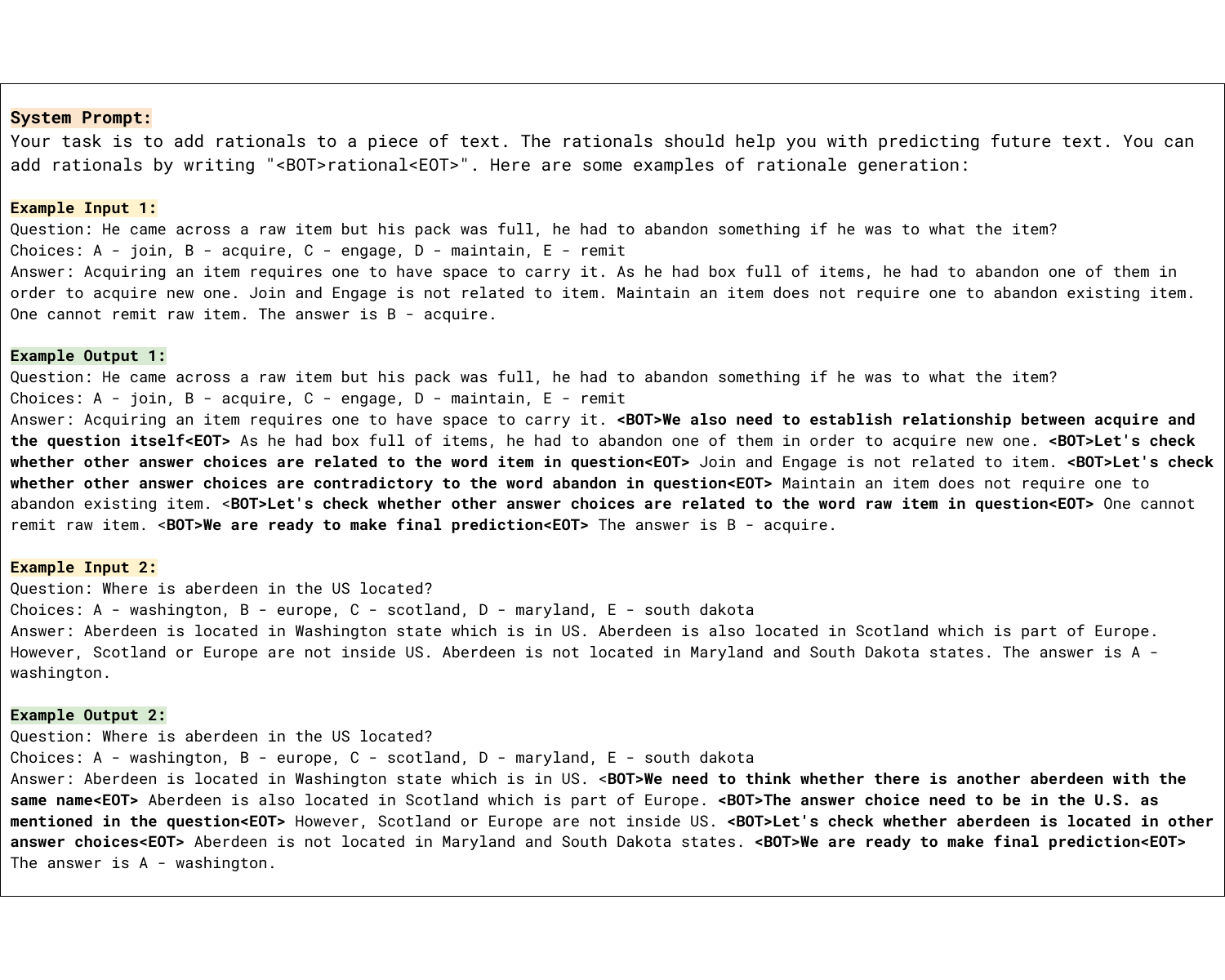} % Second figure file name
    \caption{The prompt and in-context learning examples used for sampling rationales for ECQA. The bolded rationales represent implicit rationales in the document.}
  % \caption{Two figures side by side}
  \label{fig:sampling_ecqa}
\end{figure*}

\begin{figure*}[ht]
  \centering
  % Adjust widths to fit your document
 
    \centering
    \includegraphics[trim=0.2cm 2.9cm 0.2cm 2.9cm, width=0.8\linewidth]{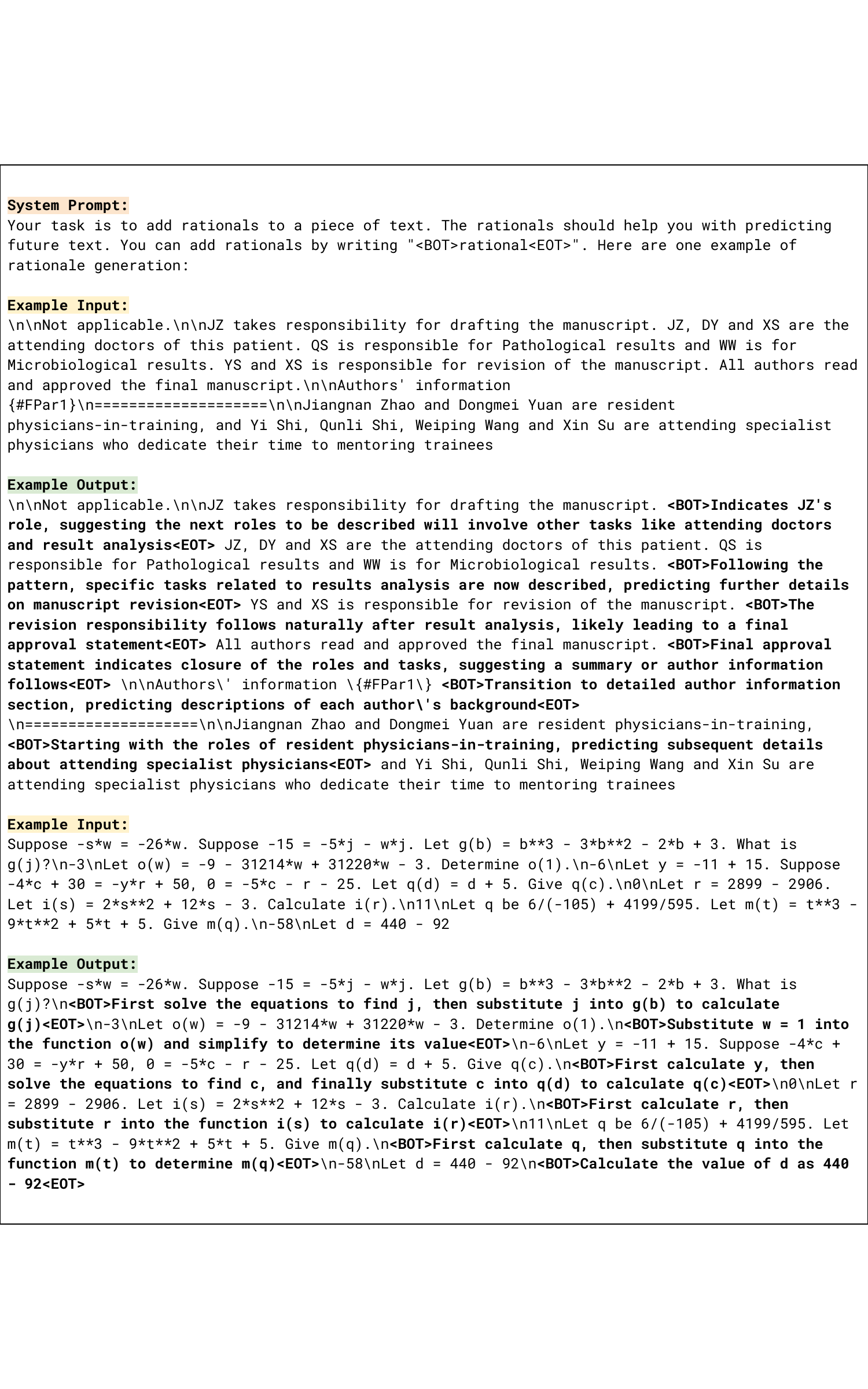} % Second figure file name
    \caption{The prompt and in-context learning examples used for sampling rationales for The Pile. The bolded rationales represent implicit rationales in the document.}
  % \caption{Two figures side by side}
  \label{fig:sampling_pile}
\end{figure*}

In this section, we provide the prompts we used for rationale sampling on GSM8K (Figure \ref{fig:sampling_gsm8k}), ECQA (Figure \ref{fig:sampling_ecqa}), and The Pile (Figure \ref{fig:sampling_pile}).

% \begin{figure*}[h]
%   \centering
%   % Adjust widths to fit your document
 
%     \centering
%     \includegraphics[trim=0.2cm 2.6cm 0.2cm 2.6cm, width=\linewidth]{figs/apps_sampling_example.pdf} % Second figure file name
%     \caption{The prompt and in-context learning examples used for sampling rationales for APPS. The bolded rationales represent model's hidden mental states during reasoning.}
%   % \caption{Two figures side by side}
%   \label{fig:sampling_apps}
% \end{figure*}

\section{Prompts used during inference}\label{app:inference_prompts}
\begin{figure*}[ht]
  \centering
  % Adjust widths to fit your document
 
    \centering
    \includegraphics[trim=0.2cm 1.3cm 0.2cm 1.3cm, width=\linewidth]{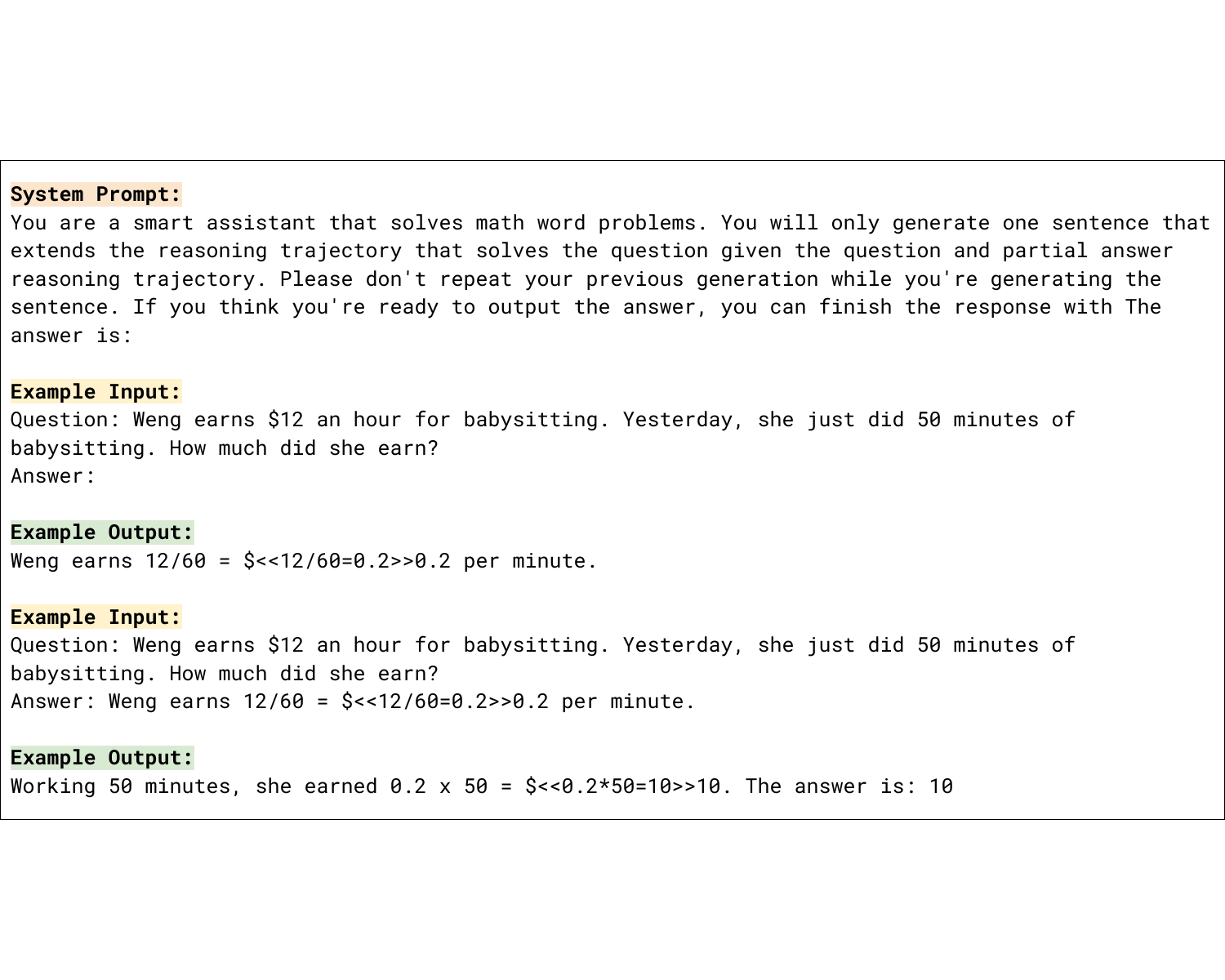} % Second figure file name
    \caption{The prompt and in-context-learning demonstrations used during inference time to encourage the agent model reason step by step on GSM8K.}
  % \caption{Two figures side by side}
  \label{fig:inference_gsm8k}
\end{figure*}

\begin{figure*}[ht]
  \centering
  % Adjust widths to fit your document
 
    \centering
    \includegraphics[trim=0.2cm 2.3cm 0.2cm 2.3cm, width=0.8\linewidth]{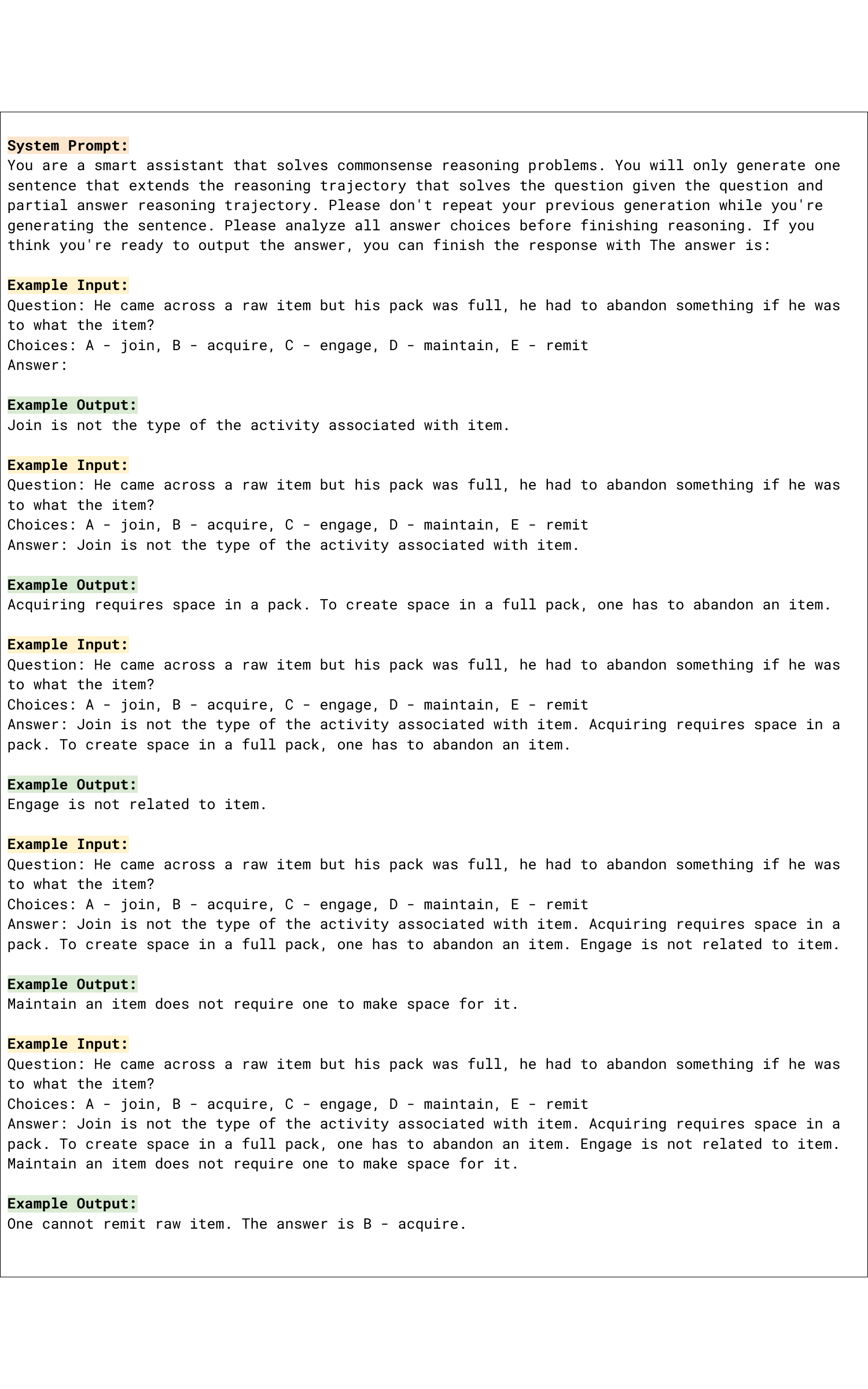} % Second figure file name
    \caption{The prompt and in-context-learning demonstrations used during inference time to encourage the agent model reason step by step on ECQA.}
  % \caption{Two figures side by side}
  \label{fig:inference_ecqa}
\end{figure*}

In this section, we provide the prompts used during inference time to encourage the agent model reason step by step for GSM8K (Figure \ref{fig:inference_gsm8k}) and ECQA (Figure \ref{fig:inference_ecqa}). Note that the input to the agent model appends the last rationale generated by the agent model.

\section{Examples of rationales generated at inference time} \label{app:example_rationale}
\begin{figure*}[ht]
  \centering
  % Adjust widths to fit your document
 
    \centering
    \includegraphics[trim=0.2cm 0.0cm 0.2cm 0.0cm, width=0.8\linewidth]{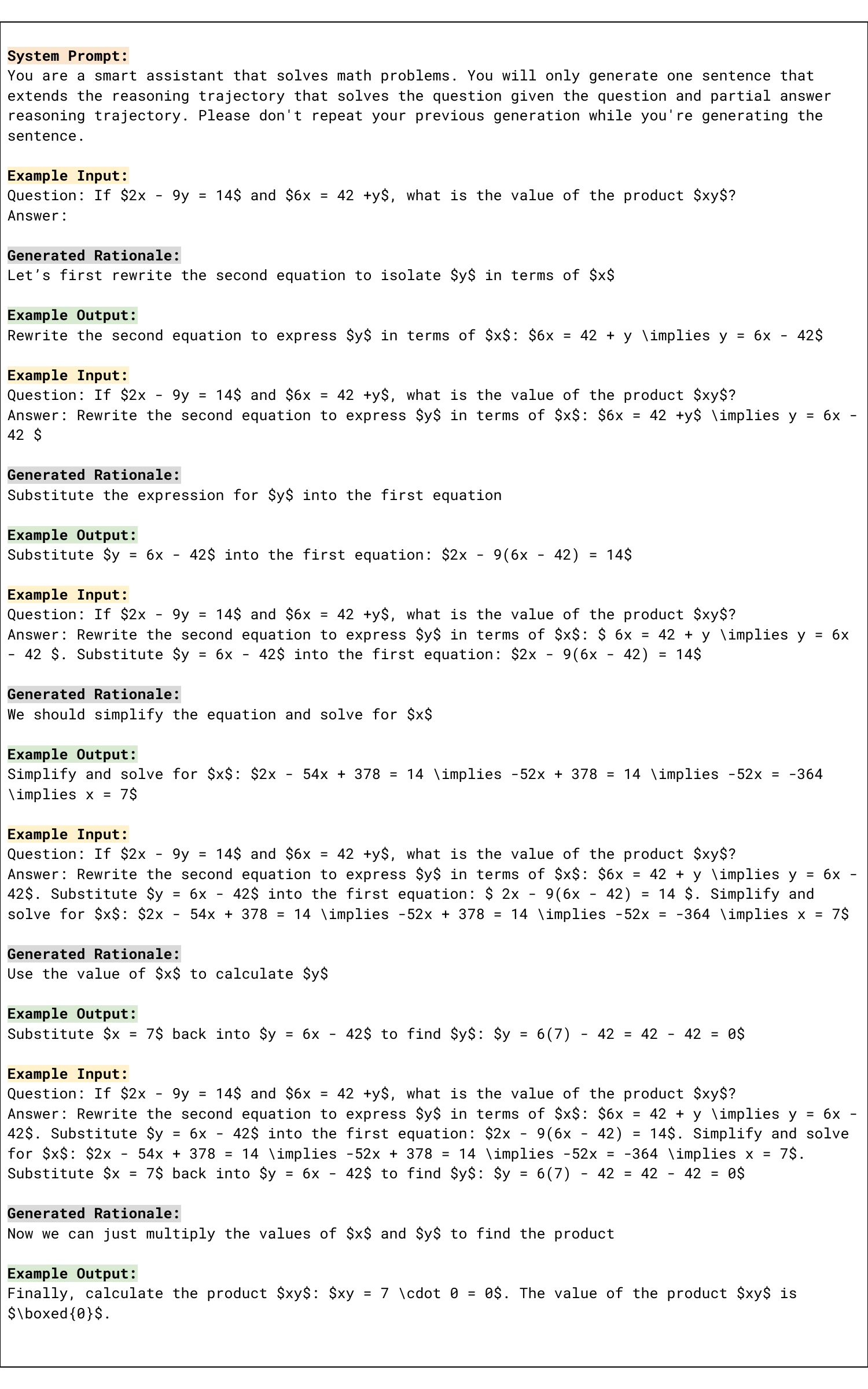} % Second figure file name
    \caption{Rationales generated by \mymethod{} for the test set of MATH.}
  % \caption{Two figures side by side}
  \label{fig:inference_math}
\end{figure*}

In this section, we provide rationales generated by \mymethod{} from the test set of MATH during inference time Figure \ref{fig:inference_math}, which was not part of the rationale sampling datasets, and observe that our model can still generate useful rationales that help to understand LLM’s reasoning process.

\section{Prompts used for LLaMa-3 reranking} \label{app:llama3-reranking}
\begin{figure*}[ht]
  \centering
  % Adjust widths to fit your document
 
    \centering
    \includegraphics[trim=0.2cm 1.3cm 0.2cm 1.3cm, width=\linewidth]{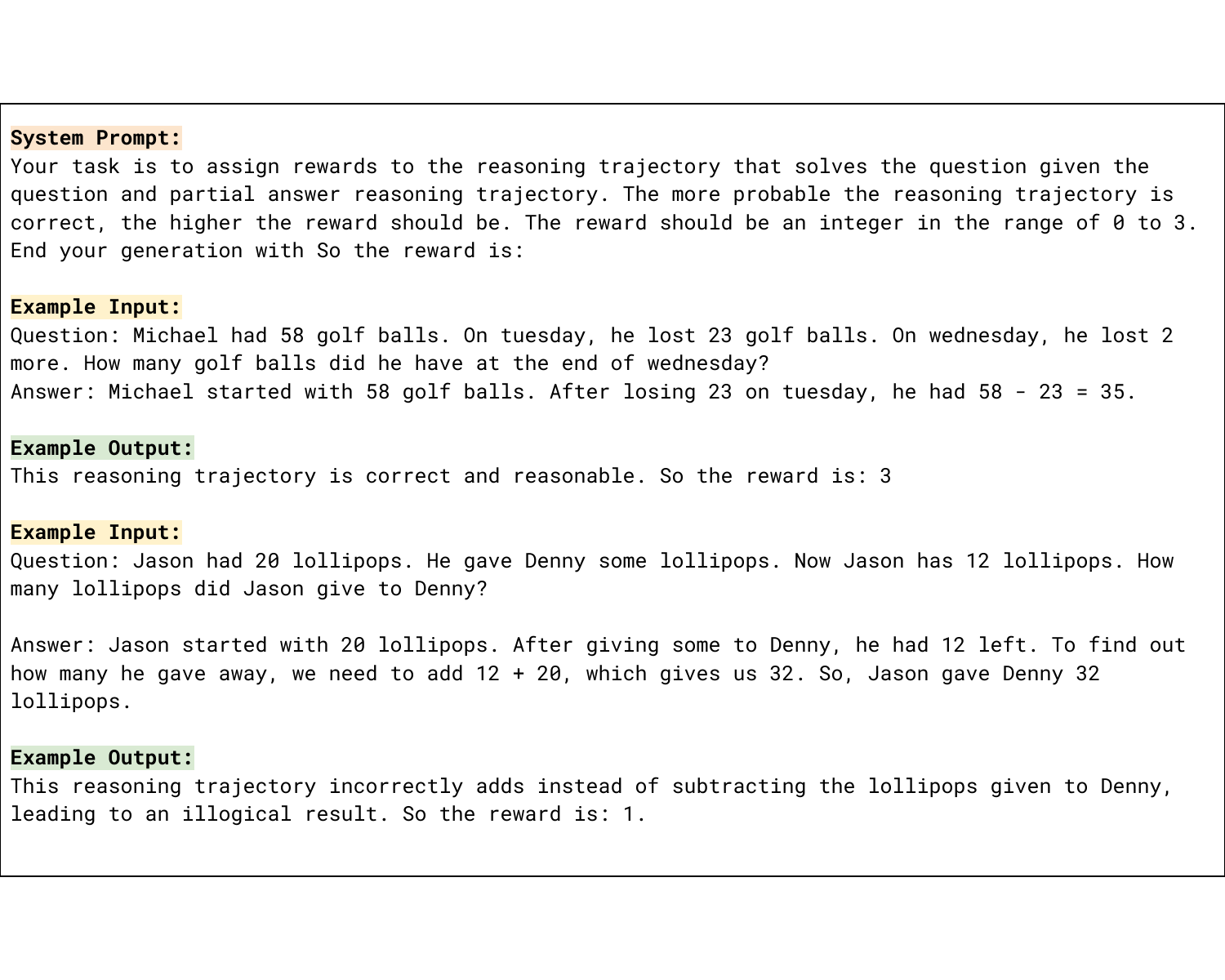} % Second figure file name
    \caption{The prompt and in-context-learning demonstrations used during process supervision to elicit the feedback by directly reranking partial reasoning trajectory.}
  % \caption{Two figures side by side}
  \label{fig:rereanking_llama3}
\end{figure*}

In this section, we provide the prompts and in-context-learning demonstrations used to instruct LLaMa-3-8B-Instruct and GPT-4 to provide feedback by directly reranking partial reasoning traces given the question (Figure \ref{fig:rereanking_llama3}).

% \paragraph{Potential improvements with advanced inference-time algorithms} Our current approach simply provides a reranking of each step of the reasoning process.
% This versatility allows our approach to be combined with inference-time search algorithms that explore reasoning trajectories to arrive at an optimal result, such as Monte-Carlo Tree Search or Tree of Thoughts   \citep{hao2023reasoning, yao2023tree}. 
% We expect this combination to augment the reasoning abilities of these approaches. 
% Other possible extensions include a ``reasoning beam-search,'' where reasoning trajectories can be sampled more effectively using beam-search with guidance from rationale perplexity. 
% Our approach is also amenable to self-consistency \citep{wang2022self}, where majority voting may be used to produce better rationales. 
% Using more sophisticated inference techniques, such as evaluating the overall perplexity of the reasoning trajectory in a manner similar to beam search, or using more sophisticated inference-time algorithms like Monto-Carlo Tree Search \citep{hao2023reasoning} might improve the final results. Our method can also benefit from self-consistency \citep{wang2022self} used by many previous works to get better rationales through the majority-voting of multiple \mymethod.

\section{Extended Related Work} \label{app:extended_related_work}
% \paragraph{Building world model.} The concept of world model has an extensive background. As described in various studies \citep{hu2023language}, world models can typically be formulated as state transition probabilities, which characterize a generative, casual mechanism of how the world state changes after an agent’s actions. 

% In the field of vision, pioneering efforts have been made in developing world models. \citet{du2023video} uses vision-language models to replicate the fundamental dynamics of the world. This world model is then used in various downstream tasks like planning  \citep{du2023video}, representation learning \citep{Garrido2024LearningAL} and robotic manipulation \citep{Bharadhwaj2024Track2ActPP}. The success of this method relies on converting diverse data formats into a uniform format \citep{yang2024learning} and utilizing generative modeling on this standardized data.

% \paragraph{Distilling hidden knowledge in pre-training corpus.} 

\paragraph{Supervising reasoning.}
Supervision-based approaches have been shown to enhance the reasoning abilities of LLMs. \citet{gsm8k} and \citet{snell2024scalingllmtesttimecompute} demonstrate that training a ``verifier" to supervise reasoning can be more parameter-efficient than simply expanding the parameters of the ``reasoner" responsible for solving the reasoning task. Ground-truth feedback from interaction with the environment is an effective form of supervision \citep{wang2023voyager}, but it works only in controlled environments like simulated world. 
Some verifiers employ principles like compositional reasoning to validate the reasoning process \citep{dhuliawala2023chainofverification, weir2024enhancing, weir2023nellie, vacareanu2024generalpurposeverificationchain}, these general-purpose approaches do not fully leverage the vast amount of available unlabeled data that a data-driven approach could utilize. Process-based supervision \citep{lightman2023lets, luo2024improvemathematicalreasoninglanguage, wang2024mathshepherdverifyreinforcellms} offers supervision at each reasoning step rather than just at the final result. While promising, it requires substantial human annotation for the correctness of intermediate steps or the existence of ground-truth answers for data collection. Our work aims to address these challenges by proposing a data-centric process-supervision method without the need for human annotation.

\paragraph{Knowledge extraction from unlabelled data.}
% next-token prediction does not always provide the complete set of reasons and often involves "leaps of logic" and hence, "next-token" prediction is not harness the full ability of LLMs to reason.
% \daniel{"potential" for what? The first sentence needs to tie to "knowledge extraction", which is the focus of this paragraph. Perhaps yo can rewrite the first 3 sentences a bit more concisely. }. 
LLMs are conventionally trained on extensive web data using autoregressive next-token prediction. While effective, this approach may not fully harness the potential of the pre-training data, as latent information within this data could be better accessed using techniques beyond simple next-token prediction.
Recent research has demonstrated several approaches to utilize this latent information to develop more sophisticated language model capabilities.
% \daniel{"potential" for what? The first sentence needs to tie to "knowledge extraction", which is the focus of this paragraph. Perhaps yo can rewrite the first 3 sentences a bit more concisely. }. 
 \citet{schick2023toolformer} introduced Toolformer, which autonomously annotates and extracts appropriate positions, names, and inputs for tool use by leveraging supervision from future tokens. Similarly, \citet{cornille2024learningplanlanguagemodeling} developed a method for learning to plan coherent article writing through self-supervised learning in text. More closely related to our work, \citet{zelikman2024quietstar} proposed Quiet-Star, which applied a comparable technique to uncover underlying rationales in daily communication to enhance reasoning capabilities.
Our work adopts a strategy similar to Quiet-Star for extracting rationales in an unsupervised manner. However, our approach diverges in its primary objective: we aim to train a ``supervisor" that can utilize these rationales to provide process supervision for any ``reasoner." This focus enables us to implement a simpler and more reliable method, as we don't need to directly integrate rationale extraction with ``reasoner" training. Our approach thus offers a novel perspective on leveraging latent information in language models to enhance their capabilities.

\paragraph{Rationales as the basis for reasoning.} 

% Rationales play a crucial role in human reasoning and its accuracy \citep{rips1994psychology, mercier2011humans}. Correct rationales typically lead to accurate reasoning outcomes, making them useful for supervising reasoning processes \citep{tversky1982judgment, davis1984diagnostic}. For LLMs, rationales are also instrumental to their reasoning capability.
Various studies have focused on improving the use of rationales to elicit reasoning. \citet{complexity} refine rationales for more effective reasoning elicitation, while \citet{li2023making} explore different approaches to leveraging rationales to enhance reasoning. Other works, such as \citet{hwang2024selfexploreavoidpitimproving}, examine the verification of rationales produced by LLMs during reasoning to improve performance. Additionally, training LLMs on rationale-rich data is a common strategy for enhancing reasoning skills. As highlighted by \citet{quantitive_llm} and \citet{jiang2024leanreasoner}, LLMs trained on science and math data tend to perform better on reasoning tasks, particularly when CoT prompting is used. In this work, we build on this foundation by using rationales as the core of our method to supervise reasoning.
%Recent studies further discovered training on rationals can enhance LLM's reasoning capability in a general way. As noted by \citet{quantitive_llm}, LLMs trained with science and math data do better on tasks that require reasoning, especially when using CoT prompting. Other results by \citet{complexity} suggest that powerful LLMs obtain advanced reasoning capabilities from being trained on code. 
% Other results by \citet{fu2022gptroadmap} and \citet{complexity} suggest that powerful LLMs gain advanced reasoning capabilities from being trained on code, as it includes many rationales in the form of comments. 

% \paragraph{Enhancing the reasoning capability of LLMs.} Reasoning is the bedrock of human intelligence. There have been various efforts to improve reasoning performance in LLMs. Some approaches focus on better prompt engineering to aid reasoning \cite{Fu2022ComplexityBasedPF, zou2024generalizablechainofthoughtpromptingmixedtask}, while others introduce topological variants \citep{yao2023tree, jiang2024respromptresidualconnectionprompting}. Additional methods include verification and refinement \citep{paul-etal-2024-refiner}, question decomposition \citep{zhou2022least}, and knowledge enhancement \citep{dhuliawala2023chainofverification}.
% Despite these advancements, LLMs continue to face challenges, including struggles with generalization and handling complex reasoning tasks \citep{greedy_reasoner, rein2023gpqagraduatelevelgoogleproofqa, nezhurina2024alicewonderlandsimpletasks}.

\section{The Resulting Data from Extraction/Filteration.} \label{resulting_data}
 On GSM8K, our method generates an average of 2.34 rationales per document, while on ECQA, it generates 2.58 rationales per document. The filtration process removes 80.5\% of the generated rationales on GSM8K and 42.4\% on ECQA.

For The Pile, we report the number of rationales per document and the number after filtration for each subdomain. The Pile's documents, being longer than those in GSM8K and ECQA, yield a higher average number of rationales per document. Among the subdomains, StackExchange retains the highest percentage of rationales, likely due to its question-answering format aligning well with our reasoning tasks and containing more inherent reasoning. However, The Pile as a whole contains less reasoning content, making rationale extraction challenging. Setting the threshold to 0 accepts all rationales more helpful than not having them, but the yield remains low. A manual review shows that most filtered rationales describe the preceding context rather than guiding future reasoning.

In total, we extracted approximately 14k rationales from GSM8K and ECQA combined, and about 65k from The Pile after filtration.

\section{\mymethod{} in Combination with General Methods for Reasoning Enhancement} \label{self-consistency}
We have conducted additional experiments focusing on GSM8K and Math datasets to verify our claim. We specifically chose these arithmetic reasoning benchmarks over multiple-choice questions, as using self-consistency on multiple-choice tasks could artificially inflate performance. Using 64 generations for self-consistency, we observed improved accuracy on both benchmarks with the help of \mymethod{}.

\end{document}